\def\arxivversion{}  %
\documentclass[10pt,twocolumn,letterpaper]{article}

\usepackage[pagenumbers]{wacv}  %

\usepackage[accsupp]{axessibility}  %

\usepackage{caption}   %
\usepackage{multirow}
\usepackage{colortbl}
\usepackage{graphicx}   %
\usepackage{tikz}       %
\usepackage{booktabs}
\usepackage{varwidth}
\usepackage{cuted}

\ifdefined\arxivversion
  \DeclareRobustCommand{\suppref}[1]{\cref{#1}}
  \DeclareRobustCommand{\mainref}[1]{\cref{#1}}
\else
  \DeclareRobustCommand{\suppref}[1]{\mbox{suppl-}\cref*{S-#1}}
  \DeclareRobustCommand{\mainref}[1]{\mbox{main-}\cref*{#1}}
\fi

\usepackage[normalem]{ulem}

\newcommand{\clamp}{\operatorname{clamp}}

\usepackage{xspace}

\definecolor{firstcolor}{RGB}{150,160,210}
\definecolor{secondcolor}{RGB}{196,216,235}
\newcommand{\first}[1]{\cellcolor{firstcolor}{#1}}
\newcommand{\second}[1]{\cellcolor{secondcolor}{#1}}

\newcommand{\cmpdir}{figures/figure_comp/figure_comparison}
\newlength{\cmpw}

\expandafter\def\csname cmpcrop@full/MipNeRF360_bonsai\endcsname{\draw[cropmark] (0.3977,0.3167) rectangle (0.7569,0.5476);}
\expandafter\def\csname cmpcrop@full/MipNeRF360_counter\endcsname{\draw[cropmark] (0.2375,0.5000) rectangle (0.6566,0.7861);}
\expandafter\def\csname cmpcrop@full/MipNeRF360_garden\endcsname{\draw[cropmark] (0.2313,0.6429) rectangle (0.6384,0.9286);}
\expandafter\def\csname cmpcrop@full/StaticHikes_forest1\endcsname{\draw[cropmark] (0.4844,0.0361) rectangle (0.9219,0.3889);}
\expandafter\def\csname cmpcrop@full/StaticHikes_forest2\endcsname{\draw[cropmark] (0.0398,0.2889) rectangle (0.3937,0.5750);}
\expandafter\def\csname cmpcrop@full/StaticHikes_university2\endcsname{\draw[cropmark] (0.4781,0.2181) rectangle (0.8320,0.5042);}
\expandafter\def\csname cmpcrop@full/TUM_rgbd_dataset_freiburg1_desk\endcsname{\draw[cropmark] (0.2344,0.2042) rectangle (0.8906,0.5417);}
\expandafter\def\csname cmpcrop@full/TUM_rgbd_dataset_freiburg2_xyz\endcsname{\draw[cropmark] (0.4625,0.4563) rectangle (0.9781,0.7688);}
\expandafter\def\csname cmpcrop@full/TUM_rgbd_dataset_freiburg3_long_office_household\endcsname{\draw[cropmark] (0.3469,0.6875) rectangle (0.8625,1.0000);}
\expandafter\def\csname cmpcrop@large_scale/SmallCityFront\endcsname{\draw[cropmark] (0.5787,0.0000) rectangle (1.0000,0.2857);}
\expandafter\def\csname cmpcrop@mip_tandt_db/db_drjohnson\endcsname{\draw[cropmark] (0.0000,0.2900) rectangle (0.4129,0.5753);}
\expandafter\def\csname cmpcrop@mip_tandt_db/db_playroom\endcsname{\draw[cropmark] (0.5854,0.6430) rectangle (1.0000,0.9291);}
\expandafter\def\csname cmpcrop@mip_tandt_db/MipNeRF360_bonsai\endcsname{\draw[cropmark] (0.2110,0.0934) rectangle (0.4792,0.2762);}
\expandafter\def\csname cmpcrop@mip_tandt_db/MipNeRF360_counter\endcsname{\draw[cropmark] (0.2246,0.3256) rectangle (0.5071,0.5183);}
\expandafter\def\csname cmpcrop@mip_tandt_db/MipNeRF360_garden\endcsname{\draw[cropmark] (0.1850,0.2857) rectangle (0.6662,0.5714);}
\expandafter\def\csname cmpcrop@mip_tandt_db/MipNeRF360U_bicycle\endcsname{\draw[cropmark] (0.4689,0.0791) rectangle (0.8868,0.3650);}
\expandafter\def\csname cmpcrop@mip_tandt_db/MipNeRF360U_bonus_flowers\endcsname{\draw[cropmark] (0.4228,0.0725) rectangle (0.8376,0.3587);}
\expandafter\def\csname cmpcrop@mip_tandt_db/MipNeRF360U_bonus_treehill\endcsname{\draw[cropmark] (0.0931,0.0000) rectangle (0.5817,0.2861);}
\expandafter\def\csname cmpcrop@mip_tandt_db/MipNeRF360U_kitchen\endcsname{\draw[cropmark] (0.2760,0.7113) rectangle (0.6739,0.9615);}
\expandafter\def\csname cmpcrop@mip_tandt_db/MipNeRF360U_room\endcsname{\draw[cropmark] (0.5806,0.4287) rectangle (1.0000,0.7148);}
\expandafter\def\csname cmpcrop@mip_tandt_db/MipNeRF360U_stump\endcsname{\draw[cropmark] (0.3317,0.0000) rectangle (0.7486,0.2861);}
\expandafter\def\csname cmpcrop@mip_tandt_db/tandt_train\endcsname{\draw[cropmark] (0.5969,0.2844) rectangle (0.9469,0.5706);}
\expandafter\def\csname cmpcrop@mip_tandt_db/tandt_truck\endcsname{\draw[cropmark] (0.0797,0.6429) rectangle (0.4300,0.9286);}

\tikzset{cropmark/.style={orange, line width=0.6pt}}

\newcommand{\cmpfull}[2]{%
  \begin{tikzpicture}
    \node[anchor=south west, inner sep=0pt, outer sep=0pt] (img)
    {\includegraphics[width=\cmpw]{\cmpdir/#1/#2_full}};
    \begin{scope}[x={(img.south east)}, y={(img.north west)}]
      \csname cmpcrop@#1\endcsname
    \end{scope}
  \end{tikzpicture}}

\newcommand{\cmpcropimg}[2]{%
  \begin{tikzpicture}
    \node[anchor=south west, inner sep=0pt, outer sep=0pt] (img)
    {\includegraphics[width=\cmpw]{\cmpdir/#1/#2_crop}};
    \draw[cropmark] ([shift={(0.3pt,0.3pt)}]img.south west)
    rectangle ([shift={(-0.3pt,-0.3pt)}]img.north east);
  \end{tikzpicture}}

\newcommand{\cmpscenefour}[1]{%
  \cmpfull{#1}{otf_nvs} &
  \cmpfull{#1}{i3dgs} &
  \cmpfull{#1}{ours} &
  \cmpfull{#1}{gt} \\
  \cmpcropimg{#1}{otf_nvs} &
  \cmpcropimg{#1}{i3dgs} &
  \cmpcropimg{#1}{ours} &
  \cmpcropimg{#1}{gt} \\[3pt]
}

\newcommand{\cmpscenefive}[1]{%
  \cmpfull{#1}{beta} &
  \cmpfull{#1}{fastergs} &
  \cmpfull{#1}{i3dgs} &
  \cmpfull{#1}{ours} &
  \cmpfull{#1}{gt} \\
  \cmpcropimg{#1}{beta} &
  \cmpcropimg{#1}{fastergs} &
  \cmpcropimg{#1}{i3dgs} &
  \cmpcropimg{#1}{ours} &
  \cmpcropimg{#1}{gt} \\[3pt]
}

\usepackage{amsmath}

\definecolor{cvprblue}{rgb}{0.21,0.49,0.74}
\usepackage[pagebackref,breaklinks,colorlinks,allcolors=cvprblue]{hyperref}

\title{Hierarchical Incremental Beta Splatting for Online Reconstruction}

\author{Linus Franke\\
\small University of Würzburg
\\
\small\textbf{\url{https://linus-franke.com/hibs}}
}

\begin{document}
\maketitle
\begin{strip}
  \centering
  \vspace{-1.3cm}
  \includegraphics[width=1.0\linewidth]{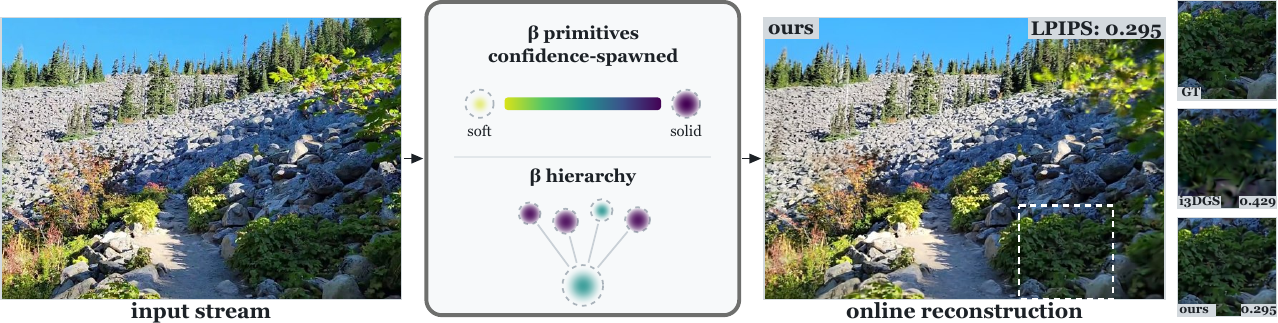}
  \vspace{-5mm}
  \captionsetup{type=figure}
  \captionof{figure}{%
    We present an online method for incremental reconstruction of primitive-based radiance fields from an unordered image stream.
    Building on beta kernels, we use kernel shape as a confidence dial: each keyframe spawns primitives with opacity and shape initialized from measured multi-view confidence.
    Short optimization bursts, fused rasterization, error-driven relocation, and closed-form LoD merging enable efficient reconstruction.
    Compared to i3DGS~\cite{meuleman2026immediate}, we achieve higher quality with about $20\%$ faster optimization.
  }\label{fig:teaser}
\end{strip}

\begin{abstract}%
    Online 3D Gaussian Splatting reconstructs a scene during capture: each keyframe spawns primitives once and refines them for only a few dozen iterations, leaving little chance to fix poor initialization or missing detail.
  Gaussians are also inherently soft, needing many overlaps for solid surfaces.

  We present Hierarchical Incremental Beta Splatting (HIBS), which brings beta kernels to incremental reconstruction.
  Their shape parameter interpolates between a soft, Gaussian-like profile and a flat-topped, near-opaque surfel, acting as a confidence dial for geometry.
  Our confidence-aware spawning uses geometric observability and multi-view depth agreement to initialize both opacity and shape, producing solid primitives where geometry is verifiable and soft, correctable primitives where it is not.
  As online densification, error-driven relocation recycles stale primitives into photometric-error hotspots.
  Bounded beta kernel support enables exact tile culling and faster rasterization, and a closed-form, shape-preserving merging formulation provides a level-of-detail (LOD) hierarchy for fast optimization and real-time rendering.
  On Mip-NeRF 360 in the on-the-fly protocol, HIBS improves PSNR by up to 0.8 dB while reducing optimization time by up to 40\%.
\end{abstract}%

\footnotetext[0]{\small\tt \{firstname.lastname\}@uni-wuerzburg.de}
\section{Introduction}
\label{sec:intro}

3D Gaussian Splatting (3DGS)~\cite{kerbl3dgs} has made high-quality radiance-field capture practical, but the standard pipeline is offline: images are collected, poses are recovered by structure-from-motion~\cite{schoenberger2016sfm}, and the scene is optimized for tens of thousands of iterations.
Recent work instead reconstructs \emph{during} capture~\cite{meuleman2025onthefly,matsuki2024monogs,keetha2024splatam}, providing immediate feedback and reducing the delay from capture to observable reconstruction to seconds.
We target this online regime by reconsidering the primitive itself.

Online reconstruction inverts the assumptions of offline 3DGS.
Each incoming image is registered as a \emph{keyframe}, spawns new primitives \emph{once}, and is optimized only for a short burst before the next keyframe arrives.
This leaves little room for gradient-driven clone/split densification, opacity resets, or long optimization schedules to repair poorly initialized primitives.
In practice, what spawning produces is largely what the model keeps.
The critical decisions---\emph{what to place and how much to trust it}---must therefore be made at initialization from measurable evidence.

We argue that under these constraints the \emph{primitive itself} can be improved, and that the Gaussian is suboptimal.
A Gaussian is fully opaque only at its center and translucent elsewhere; representing a solid surface therefore requires overlapping many Gaussians, together with the densification and long optimization that offline 3DGS can afford.
Without this machinery, online reconstruction must spawn translucent Gaussians and hope a short optimization burst can arrange them correctly.

We instead use the \emph{beta kernel}~\cite{liu2025dbs}, whose bounded profile has a learnable shape parameter $b$ that continuously interpolates between a Gaussian-like volume and a flat-topped, near-opaque surfel.
A single beta primitive can thus be a flat-topped surface patch while retaining a soft representation for uncertain geometry, thin structures, or volume.

While Deformable Beta Splatting~\cite{liu2025dbs} uses this parameter offline as a compactness degree of freedom learned over long optimization schedules, we use it from geometric evidence as a \emph{confidence axis}.
The key question is therefore not only how to optimize the shape, but whether the available evidence justifies treating a primitive as a solid surface.

That evidence must be geometric, not merely photometric.
Naively spawning solid kernels is harmful: a compact, near-opaque primitive at the wrong depth blocks transmittance and suppresses gradients behind it.
Conversely, a low photometric error does not guarantee correct geometry.
When the triangulation baseline is small relative to depth, many depths explain the observed views similarly well, yet only a narrow range remains consistent under parallax.

We therefore use the beta shape parameter as a \emph{confidence dial}.
Confidence-aware spawning combines two independent geometric cues, Multi-view stereo (MVS) agreement with an aligned monocular prior and depth observability, to initialize each primitive along this soft-to-solid axis: well-constrained candidates are spawned solid, while ambiguous ones remain soft and correctable.
As additional views arrive, a lifetime controller updates the same parameter from accumulated multi-view evidence, progressively turning confirmed geometric confidence into solidity.

To compensate for the absence of offline densification, we introduce \emph{error-driven relocation}.
Near-dead primitives are recycled into photometric-error hotspots, where render-preserving copies provide additional capacity precisely where the current model is deficient.

The beta representation is also performant.
Its bounded support yields an exact tile-culling bound and faster rasterization, while its closed-form moments enable a shape-aware level-of-detail hierarchy for large captures.

Together, these components yield an online reconstruction system in which the primitive is adapted to the short optimization horizon rather than relying on offline densification to compensate for it.
Our experiments show improved reconstruction quality and efficiency across online, benchmark, and large-scale captures.

In summary, our contributions are:
\begin{itemize}
  \item We identify the Gaussian primitive as suboptimal in the \emph{spawn-once, optimize-briefly} regime and adopt the beta kernel as a \emph{confidence-adaptive} online primitive whose shape ranges from soft to solid.
  \item We introduce \emph{confidence-aware spawning}, which initializes opacity and shape from independent depth agreement and geometric observability.
  \item We introduce \emph{error-driven relocation}, a render-preserving online alternative to clone/split densification that recycles near-dead primitives into photometric-error hotspots.
  \item We derive a \emph{closed-form merge} for beta primitives, preserving their shape statistics and enabling a level-of-detail hierarchy for large-scale reconstruction.
  \item We develop a \emph{fast beta rasterizer} with a shape-correct culling bound that exploits the kernel's compact support.
\end{itemize}
\vspace{1.5mm}
\noindent The project page and source code can be found at: \mbox{\url{https://linus-franke.com/hibs}}.
\vspace{1mm}

\section{Related Work}
\label{sec:related}

\begin{figure*}[t]
  \centering
  \includegraphics[width=.9\linewidth]{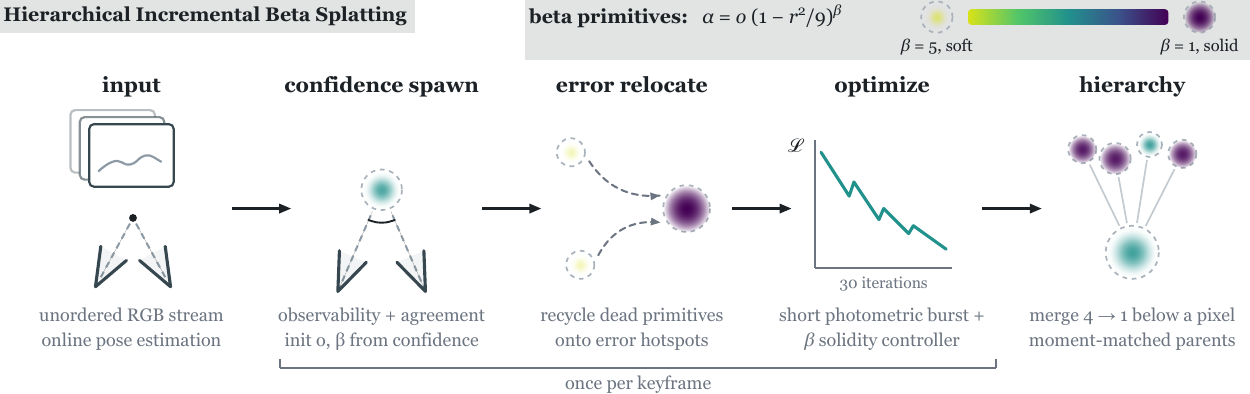}
  \vspace{-2mm}
  \caption{
    \textbf{Overview of our pipeline.}
    An RGB stream is registered with camera parameters estimated online~\cite{meuleman2026immediate}.
    Each keyframe spawns beta primitives using geometric observability and depth agreement, keeping uncertain primitives soft.
    Near-dead primitives are relocated to high-error regions, followed by a short 30-iteration optimization burst with our fused rasterizer.
    Periodically, small-footprint primitives are merged hierarchically using closed-form moment matching that preserves their solidity.
  } %
  \label{fig:pipeline}
\end{figure*}

\paragraph{Novel View Synthesis.}
Neural Radiance Fields~\cite{mildenhall2020nerf,barron2022mipnerf360} synthesize photorealistic novel views from posed images, but volumetric rendering remains computationally expensive.
3D Gaussian Splatting (3DGS)~\cite{kerbl3dgs} instead enables efficient rasterization and fast optimization.
Subsequent work improves compositing and rendering fidelity~\cite{yu2024mip,radl2024stopthepop,hahlbohm2025htgs} and scales to large scenes with hierarchical representations such as Hierarchical 3DGS~\cite{kerbl2024hierarchical} and Octree-GS~\cite{ren2024octree}.
These methods are predominantly offline, relying on precomputed poses and long optimization schedules, whereas we target reconstruction during capture.

A range of systems reconstruct 3DGS maps incrementally from video, including SLAM-based methods~\cite{matsuki2024monogs,keetha2024splatam,huang2024photoslam,peng2024rtgslam,zhu2025loopsplat,homeyer2024droid,cheng2025s3pogs} and joint pose-and-reconstruction approaches~\cite{Fu_2024_CVPR,lin2025longsplat}.
SLAM methods spend far more per keyframe, e.g., hundreds of mapping iterations, densification, and long final refinement~\cite{matsuki2024monogs,zhu2025loopsplat}.
Feed-forward and hybrid methods~\cite{jiang2025anysplat,lin2025depthanything3,li2025streamgs,li2026artdeco} trade scene-specific optimization for speed, typically at reduced fidelity.
Closest to our setting, On-the-Fly NVS~\cite{meuleman2025onthefly} and i3DGS~\cite{meuleman2026immediate} reconstruct image streams with short optimization bursts and no densification, using learned matching and hierarchical structures.
However, these methods retain the Gaussian and spawn it once per view.
Our work instead revisits the primitive itself, introducing a controllable notion of solidity specifically for this short-burst regime.

\paragraph{Primitive Kernels and Densification.}
Alternative splatting primitives have been proposed to improve compactness, surface accuracy, or geometric fidelity, including generalized and radial kernels~\cite{hamdi2024ges, huang2025deformable}, planar Gaussians~\cite{huang20242dgs}, point sprites~\cite{ruckert2022adop,aliev2020neural, harrer2023inovis}, linear and convex primitives~\cite{vonluetzow2025linprim,held2025convex}, volumetric ellipsoids~\cite{mai2025ever}, and trilinear splats~\cite{franke2024trips}.
Deformable radial kernels~\cite{huang2025deformable} are flexible, but planar and without closed-form moments, which our spawning and merge rely on.
Most closely related, Deformable Beta Splatting (DBS)~\cite{liu2025dbs,liu2026universal} introduces a bounded beta kernel with a learnable shape parameter spanning Gaussian-like and surfel-like profiles.
DBS learns this shape offline together with densification; we instead initialize it from geometric evidence and update it online throughout.

3DGS densification has been refined through error- and gradient-based splitting, budget-aware pruning, and MCMC-based relocation~\cite{rotabulo2024revising,yu2024gaussian,fang2024minisplatting,kheradmand2024mcmc}.
In particular, 3DGS-MCMC~\cite{kheradmand2024mcmc} reuses low-opacity primitives by relocating them onto live ones with render-preserving corrections.
These techniques assume long offline optimization and periodic densification, whereas our online setting has neither.
We adapt relocation to this regime by recycling near-dead primitives and targeting them using attributed photometric error.

\paragraph{Efficient Rasterization.} %
Recent work accelerates 3DGS through improved backward passes, gradient reduction, tighter tile bounds, and fused optimization~\cite{mallick2024taming,durvasula2024distwar,radl2024stopthepop,hanson2025speedy,hahlbohm2026fastergs}.
We build on these ideas for the online burst, but derive the rasterization bounds for the bounded beta kernel.
In particular, the Gaussian cutoff is not valid for low-opacity solid kernels, making the beta-specific bound a correctness requirement in addition to an efficiency improvement.

\paragraph{Depth Priors and Uncertainty.}
Monocular and multi-view depth predictors are commonly used to initialize or regularize radiance-field reconstruction~\cite{yang2024depthanythingv2,lin2025depthanything3,wang2025vggt}; for example, Meuleman \etal.~\cite{meuleman2025onthefly} combine aligned monocular depth with guided MVS.
Uncertainty is often estimated post-hoc or used for view selection~\cite{goli2024bayesrays,jiang2024fisherrf}.
We instead incorporate uncertainty directly into the online primitive lifecycle: agreement between independent depth estimates and geometric observability determine the primitive's opacity and shape, while the evolving shape field provides a live confidence representation.

\section{Preliminaries: Online 3DGS}
\label{sec:prelim}

Online 3DGS systems~\cite{meuleman2025onthefly,li2026artdeco} reconstruct an RGB stream incrementally: each incoming frame is registered, used to spawn new primitives, and followed by a short optimization burst before the next keyframe.
In i3DGS~\cite{meuleman2026immediate}, registration matches each frame against previously registered keyframes using place recognition and learned feature matching~\cite{alibey2023mixvpr,potje2024xfeat,lindenberger2023lightglue}, initializes the pose with PnP/RANSAC, and performs local bundle adjustment over a covisibility graph, all running in tens of milliseconds per frame.

After registration, pixels with local image detail are sampled and lifted to 3D using available multi-view and monocular depth cues~\cite{meuleman2025onthefly,meuleman2026immediate,li2026artdeco}.
The samples are instantiated as low-opacity, translucent isotropic Gaussians, followed by a short optimization burst of only a few dozen iterations.
Crucially, these online systems do not perform the clone/split densification or opacity resets used by offline 3DGS, nor the long per-keyframe mapping and final refinement of Gaussian SLAM~\cite{matsuki2024monogs}: primitives are spawned once and then retained, making their initialization critical.

\section{Method}

Our method reconstructs a primitive-based radiance field online from an unordered stream of RGB images.
We adopt the pose-estimation backbone of Meuleman et al.~\cite{meuleman2026immediate}, but redesign the primitive and its lifecycle.
In the spawn-per-view, short-burst regime, positions receive too little optimization to recover from poor initialization, while longer optimization overfits already-seen keyframes.
We therefore make confidence explicit in the primitive representation: each primitive can be either a soft probe or a self-supporting surface patch, and primitives are later relocated from unused capacity to image-space error hotspots.
An overview of our method can be found in \cref{fig:pipeline}.

\subsection{The Beta Kernel as a Confidence Dial} \label{sec:beta_prim}

Following Deformable Beta Splatting~\cite{liu2025dbs}, we replace the Gaussian falloff with a bounded beta kernel.
With $x=r^2/9$ in the primitive's local frame (support $r \le 3$, using the common $3\sigma$ extent), a primitive of opacity $o$ and shape parameter $b\in\mathbb{R}$ contributes
\begin{equation}
  \alpha = o\,\mathcal{B}(x;b),\qquad
  \mathcal{B}(x;b)=(1-x)^{\beta(b)},\qquad
  \beta(b)=4e^b .
  \label{eq:beta_kernel}
\end{equation}
At $b=0$ ($\beta=4$) the profile closely matches a Gaussian; decreasing $\beta$ flattens it toward a solid plateau, while increasing $\beta$ produces a soft, rapidly vanishing blob.
Thus opacity $o$ controls contribution strength and $\beta$ controls the primitive's degree of geometric commitment.

We also use the spherical-beta color basis of DBS,
\begin{equation}
  \mathbf C(\mathbf v)=\mathbf c_0+\sum_k \mathbf a_k\, \max(0,\mathbf d_k\!
  \cdot\!\mathbf v)^{4e^{b_k}}.
  \label{eq:sph_beta}
\end{equation}
Unlike offline DBS, where nonzero lobe amplitude initializations are used and optimized away over thousands of iterations, online primitives receive only a few dozen updates.
We therefore initialize all lobe amplitudes at zero with random orientations.
This removes the persistent view-dependent color bias of the offline initialization, and works as the amplitude gradient remains nonzero at zero.

\subsection{Observability-Aware Spawning}
\label{sec:spawn}

\begin{figure}
  \includegraphics[width=\linewidth]{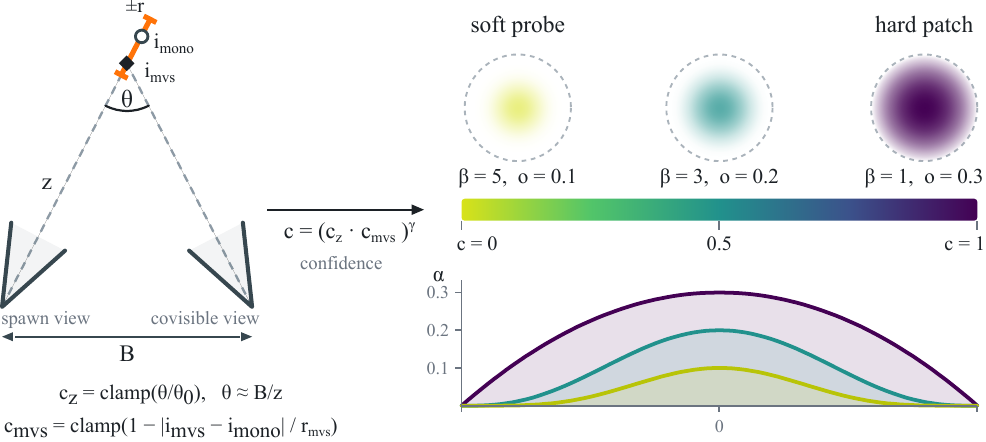}
  \vspace{-2mm}
  \caption{\textbf{Spawning/parameter initialization.}
    Based on depth observability $c_{\mathrm z}$ and agreement of depths (orange window) $c_{\mathrm{mvs}}$, the beta primitives are initialized as hard patches or soft probes.
  } %
  \label{fig:spawn} %
\end{figure}

Each registered keyframe spawns primitives where the current model under-explains image detail.
With the short optimization burst following and no guarantee of repeated observations, opacity and shape must be initialized well from observable geometry.
We combine geometric depth observability with cross-validation between two independent depth estimators, as also depicted in \cref{fig:spawn}.

\paragraph{Depth observability.
}
For a maximum covisible baseline $B$ and candidate depth $z$, the triangulation angle is approximately $B/z$.
We therefore use
\begin{equation}
  c_{\mathrm z}=
  \clamp\!\left(\frac{B/z}{\theta_0},0,1\right),
  \qquad \theta_0=2^\circ .
\end{equation}
Large $c_{\mathrm z}$ indicates that depth is geometrically constrained; small $c_{\mathrm z}$ indicates that many depths explain the training views: the estimate is unreliable under novel-view parallax.

\paragraph{Confidence-aware spawning.}
Pixel $\mathbf p$ is sampled according to image detail of image $I$ not already explained by the render of the current model $R$, following~\cite{meuleman2025onthefly},
\begin{equation}
  p(\mathbf p)=
  \max\!\Big(\phi\, s_I(\mathbf p),p_{\min}\Big) - \phi\, s_R(\mathbf p),
  \label{eq:zipprob}
\end{equation}
where $s_I,s_R$ are blurred-Laplacian magnitudes, $p_{\min}$ the lower probability bound, and the scalar $\phi$ the base sampling scale that sets the spawn density (\suppref{tab:suppl-hparams}).

Candidate depths $i_\mathrm{mvs}$ are obtained with guided MVS in an inverse-depth window of radius $r_{\mathrm{mvs}}$ around the metrically aligned monocular prior $i_\mathrm{mono}$.
We introduce their agreement as an independent confidence measure:
\begin{equation}
  c_{\mathrm{mvs}}=
  \clamp\!\big(1-|i_{\mathrm{mvs}}-i_{\mathrm{mono}}|/r_{\mathrm{mvs}},0,1\big),
  \label{eq:costconf}
\end{equation}
set to zero for flat matching-cost profiles. We combine the two signals as $c=(c_{\mathrm{mvs}} \cdot c_{\mathrm z})^\gamma$ and initialize
\begin{equation}
  \begin{aligned}
    o_{\mathrm{spawn}}     & =o_0+(o_1-o_0)c,       \\
    \beta_{\mathrm{spawn}} & =\beta_{\mathrm{soft}}
    -(\beta_{\mathrm{soft}}-\beta_{\mathrm{solid}})c,
  \end{aligned}
\end{equation}
with $\gamma=4$, $o_0=0.1$, $o_1=0.3$, $\beta_{\mathrm{solid}}=1$, and $\beta_{\mathrm{soft}}=5$.
When MVS provides no reliable correspondence ($c_{\mathrm{mvs}}=0$), we use the conservative opacity floor $o_{\mathrm{floor}}=0.02$.

The primitive scale is calibrated for $\beta=4$ and compensated to preserve the half-maximum footprint:
\begin{equation}
  s\leftarrow s
  \sqrt{\frac{u_{1/2}(\beta_1)}{u_{1/2}(\beta_2)}},
  \qquad
  u_{1/2}(\beta)=1-2^{-1/\beta},
  \label{eq:scale_compensation}
\end{equation}
with $\beta_1=4$ and $\beta_2=\beta_{\mathrm{spawn}}$.

\subsection{Controlling Primitive Solidity}
\label{sec:solidity}

The spawn decision is extended over the primitive's lifetime by a deterministic controller applied once per keyframe.
For each visible primitive, we track exponential moving averages ($\eta=0.3$) of the mean $\bar e$ and variance $\sigma_e^2$ of the photometric error over the pixels it dominates, across views.
The shape parameter is updated as \[ b\leftarrow b\pm\Delta b,\qquad \Delta b=0.4,\qquad \beta=4e^b, \] with $\beta$ decreased (hardened) when the primitive is old ($\geq$10 keyframes), opaque ($o>0.7$), well observed ($>$20 dominated pixels, statistics from $\geq$3 views), and has low mean error ($\bar e<0.03$), and increased (softened) toward $\beta=5$ only when well observed high error is strongly view-inconsistent ($\bar e>0.03$, $\bar e^2/(\bar e^2{+}\sigma_e^2)<0.5$).
View-consistent error is left to relocation (see next section), since it indicates geometric error rather than view-dependent appearance.
Updates are clamped to $\beta\in[1,5]$ and use the scale compensation from \cref{eq:scale_compensation} with $\beta_1=\beta_{\mathrm{old}}$ and $\beta_2=\beta_{\mathrm{new}}$.
Note that through gradient descent, $\beta$ is free to be optimized outside the clamped range.

\begin{figure}
  \includegraphics[width=\linewidth]{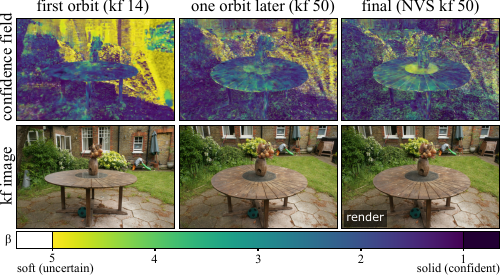}
  \vspace{-6mm}
  \caption{\textbf{Online confidence field.}
    After the first orbit (left), many background primitives remain uncertain, and the table reconstruction is fuzzy.
    After the second orbit (middle), accumulated multi-view evidence solidifies most primitives.
    In the final reconstruction (right, novel view rendering), these areas are largely solidified, while uncertainty remains around the reflective table center.
  } \label{fig:conf_field}%
\end{figure}

Thus $\beta$ acts as an evidence-driven confidence state: low $\beta$ denotes earned geometric commitment, while high $\beta$ denotes unresolved or view-dependent evidence.
The resulting shape field provides an immediate confidence map of the reconstruction, as illustrated in \cref{fig:conf_field}.

\subsection{Error-Driven Incremental Relocation} \label{sec:error_reloc}

The overall amount of new detail that can be added per keyframe is limited, and offline clone/split heuristics require longer optimization.
Instead, inspired by the relocation mechanism of 3DGS-MCMC~\cite{kheradmand2024mcmc}, we introduce an error-targeted split: near-dead primitives are recycled by teleporting them to poorly reconstructed live primitives.

Once per keyframe, dead primitives, being below the recycling-opacity threshold, outside the hierarchy, and past the spawn-protection window, serve as sources.
Live destinations are sampled according to their per-primitive error attribution, weighted by error consistency and restricted to the top $10\%$, so high-error regions receive more capacity.
Each selected source is teleported to be a copy of the destination's position, shape, and color, producing stacks of relocated primitives whose size grows with residual error.

Before optimization, copy stacks are made render preserving, by adjusting their combined opacity and scale~\cite{kheradmand2024mcmc}.
Small positional jitter and optimizer resets allow the subsequent burst to separate the copies and represent more detail.

\subsection{Per-keyframe Optimization Burst}
\label{sec:optim}

Each burst minimizes the photometric objective~\cite{kerbl3dgs},
\begin{equation}
  \mathcal L=(1-\lambda)\mathcal L_1+
  \lambda(1-\mathcal{L}_\mathrm{SSIM}) +\lambda_{\mathcal{R}}\mathcal R_{\mathrm{o}} + \lambda_{\mathcal{R}}\mathcal R_{\mathrm{\Sigma}},
  \quad \lambda=0.15,
\end{equation}
with learned exposure/ISP~\cite{deutsch2026ppispphysicallyplausiblecompensationcontrol} correction applied before comparison.
We additionally use the opacity ($\mathcal R_{\mathrm{o}}$) and scale ($\mathcal R_{\Sigma}$) regularizers of 3DGS-MCMC~\cite{kheradmand2024mcmc}, also used by DBS~\cite{liu2025dbs}, both with weight $\lambda_{\mathcal{R}}=10^{-2}$.
The opacity penalty drives unsuccessful primitives toward the recycling band rather than pruning them outright.

\subsection{Fast Beta Rasterization}
\label{sec:raster}

The bounded support of the beta kernel permits tighter rasterization bounds than the Gaussian.
For opacity $o$, the normalized squared radius at the rasterizer's $1/255$ cutoff is $x_{\mathrm{cut}}= \max(0,1-(255o)^{-1/\beta(b)})$.
We use this bound for tile culling.
The standard 3DGS Gaussian threshold is insufficient for solid, low-opacity beta kernels and can discard tiles containing nonzero contributions; the beta-specific bound is therefore both a correctness fix and a speedup.

The optimization burst is further accelerated with a FasterGS-style fused rasterizer~\cite{hahlbohm2026fastergs}: backward recomputes forward quantities per primitive, keeps intermediate gradients in registers, and applies sparse Adam directly to the optimizer state.
Primitive gradients therefore never materialize in memory.
The regularizers are also applied analytically inside the fused backward/optimizer update.
This rasterizer substantially reduces computation time.

\section{Hierarchical Beta Primitives}
\label{sec:hierarchy}
\begin{figure}
  \centering
  \includegraphics[width=.9\linewidth]{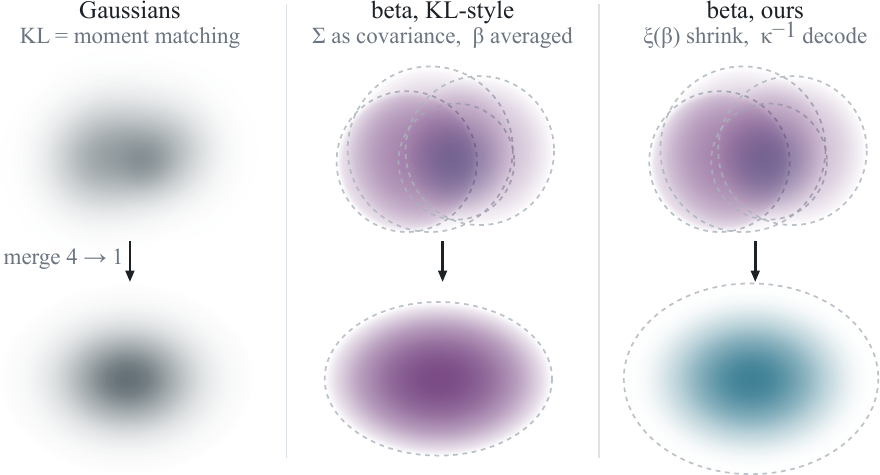}
  \vspace{-2mm}
  \caption{\textbf{Merging beta primitives}.
    KL-style merging of beta primitives (middle) similar to merging of Gaussians (left) leads to oversized and solid primitives.
    Instead, we use moment merging (right), keeping the footprint.
  }%
  \label{fig:hierarchy_merging}
\end{figure}
\begin{table*}[!ht]
  \setlength{\tabcolsep}{3pt}\centering
  \caption{\label{tab:slam-eval-s3po}
    \textbf{Results for novel view synthesis on pose-free methods.}
    The {\colorbox{firstcolor}{best}} and {\colorbox{secondcolor}{second best}} are color coded for pose-free methods.
    Times for G+Beta-Splatting and G+Faster-GS (offline baselines at 7k as a reference) include GLOMAP pose estimation (about 50s for TUM, 174s for MipNeRF360 (\textit{otf}) and 58min for StaticHikes).
    $^\dagger$ numbers taken from i3DGS~\cite{meuleman2026immediate}, with times roughly scaled to our setup.}
  \vspace{-9pt}
  \footnotesize
  \begin{tabular}{l|cccr|cccr|cccr}
    \toprule
                                                            &
    \multicolumn{4}{c|}{\textsc{TUM}}                       &
    \multicolumn{4}{c|}{\textsc{MipNeRF360} \textit{(otf)}} &
    \multicolumn{4}{c}{\textsc{StaticHikes}}                                                                                                                                                                                                                                                                                                                                                                                                                 \\
    {}                                                      & PSNR$^\uparrow$               & SSIM$^\uparrow$                & LPIPS$^\downarrow$              & Time$^\downarrow$              &
    PSNR$^\uparrow$                                         & SSIM$^\uparrow$               & LPIPS$^\downarrow$             & Time$^\downarrow$               &
    PSNR$^\uparrow$                                         & SSIM$^\uparrow$               & LPIPS$^\downarrow$             & Time$^\downarrow$                                                                                                                                                                                                                                                                                                             \\
    \midrule

    G+Beta-Splatting (7k)                                   &
    {22.73}                                                 & {0.842}                       & {0.214}                        & {0:01:50}                       & {27.75}                        & {0.863}                        & {0.232}                        & {0:04:34}                        & {20.36}                        & {0.538}                        & {0.470}                        & {0:59:22}                                          \\
    G+Faster-GS (7k)                                        &
    {23.08}                                                 & {0.845}                       & {0.207}                        & {0:01:42}                       & {27.81}                        & {0.876}                        & {0.210}                        & {0:04:22}                        & {20.57}                        & {0.547}                        & {0.459}                        & {0:59:38}                                          \\
    \hline
    \hline
    MonoGS$^\dagger$                                        &
    {16.60}                                                 & {0.682}                       & {0.381}                        & $\sim$15min                     & {14.46}                        & {0.436}                        & {0.663}                        & $\sim$4min                       & {15.46}                        & {0.301}                        & {0.659}                        & $\sim$9min                                         \\
    S3PO-GS with refinement$^\dagger$                       &
    {21.88}                                                 & {0.777}                       & {0.309}                        & $\sim$1h                        & {19.82}                        & {0.553}                        & {0.576}                        & $\sim$20min                      & {18.09}                        & {0.368}                        & {0.611}                        & $\sim$50min                                        \\
    LongSplat$^\dagger$                                     &
    \cellcolor{firstcolor}{26.51}                           & \cellcolor{firstcolor}{0.875} & \cellcolor{secondcolor}{0.176} & $\sim$1.5h                      & {23.63}                        & {0.657}                        & {0.415}                        & $\sim$2h                         & {20.05}                        & {0.461}                        & {0.481}                        & $\sim$20h                                          \\
    On-The-Fly NVS                                          &
    {23.37}                                                 & {0.824}                       & {0.241}                        & \cellcolor{firstcolor}{0:01:00} & {22.51}                        & {0.671}                        & {0.370}                        & \cellcolor{secondcolor}{0:01:45} & {21.14}                        & {0.619}                        & {0.338}                        & \cellcolor{secondcolor}{0:02:08}                   \\
    i3DGS                                                   &
    {25.36}                                                 & {0.858}                       & {0.188}                        & \second{}0:01:24                & \cellcolor{secondcolor}{26.66} & \cellcolor{secondcolor}{0.835} & \cellcolor{secondcolor}{0.224} & {0:02:09}                        & \cellcolor{secondcolor}{22.17} & \cellcolor{secondcolor}{0.699} & \cellcolor{secondcolor}{0.269} & {0:02:53}                                          \\ \hline
    Ours                                                    & \second{}25.94                & \second{}0.868                 & \first{}0.175                   & 0:01:44                        & \first{} 27.47                 & \first{}0.852                  & \first{}0.204                    & \first{}0:01:18                & \first{}22.49                  & \first{}0.710                  & \first{} 0.259                   & \first{}0:02:02 \\
    \bottomrule
  \end{tabular}
  \vspace{-3mm}
\end{table*}

Kilometre-scale captures exceed VRAM, so we maintain a screen-space level-of-detail hierarchy~\cite{meuleman2026immediate,kerbl2024hierarchical}.
Groups of four neighbouring primitives that each project below half a pixel are replaced by a coarse parent; inactive fine levels are offloaded to host memory.
For sub-pixel groups, the parent only needs to preserve the integrated footprint, which we approximate by conserving 3D mass.
Unlike a Gaussian, however, our primitive also carries shape, so the merge must preserve solidity (\cref{fig:hierarchy_merging}).
\paragraph{Closed-form merging by moment matching.}
Each primitive carries an opacity $o$, center $\mu$, shape matrix $\Sigma$, and scalar shape parameter $b$, with $\beta=4e^b$.
Its compactly supported kernel is $\rho(\mathbf{x}) = o\,(1-\|\mathbf{w}\|^2/9)^\beta$, where $\mathbf{w}$ is the whitened offset.

Two properties make closed-form merging possible.
First, the mass $Z$, covariance $\mathrm{Cov}=\xi(\beta)\Sigma$ with shrink factor $\xi(\beta)=4.5/(\beta+2.5)$, and Mardia kurtosis $\kappa(\beta)=15(\beta+2.5)/(\beta+3.5)$ all have closed forms (\suppref{sec:supp-merge-derivation}).
Here $\Sigma$ parameterizes the support ellipsoid rather than the spread, since peakier profiles concentrate mass near the center.
Second, $\kappa$ is strictly monotone in $\beta$ with $15-\kappa(\beta)=15/(\beta+3.5)$, so on the band $(75/7,15)$ it has the closed-form inverse $\beta=15/(15-\kappa)-3.5$ and uniquely identifies the profile within the family.

We merge primitives by moment matching rather than KL minimization.
For compactly supported kernels, KL is unsuitable because any nonzero mass outside the parent's support makes the divergence infinite.

With mass weights $w_i\propto Z_i$, offsets $\delta_i=\mu_i-\mu^\star$, and mixture kurtosis $\kappa_{\mathrm{mix}}$ (a closed-form mass-weighted sum of per-child trace terms; see \suppref{sec:supp-merge-derivation}), the parent is
\begin{align}
  \mu^\star   & = \sum_i w_i\,\mu_i,
  \qquad
  \Sigma_{\mathrm{mix}} = \sum_i w_i\big(\xi(\beta_i)\,\Sigma_i+\delta_i\delta_i^\top\big), \\
  \beta^\star & = \frac{15}{15-\hat\kappa}-3.5,
  \quad
  \Sigma^\star=\frac{\Sigma_{\mathrm{mix}}}{\xi(\beta^\star)},
  \quad
  o^\star=\frac{\sum_i Z_i}{Z(\beta^\star,\Sigma^\star)},
\end{align}
with $b^\star=\ln(\beta^\star/4)$ and $\hat\kappa$ the mixture kurtosis $\kappa_{\mathrm{mix}}$ clipped to the representable band of the beta family (\suppref{sec:supp-merge-derivation}).
The covariance follows from the law of total covariance, dividing out the shrink factor turns it back into a shape matrix, and mass conservation determines the opacity.

\paragraph{Large-scale training.}
The hierarchy is rebuilt periodically rather than per keyframe, with dynamic-object and sky masks applied when available.
To prevent redundant spawning, views observing primitives below one-quarter of their finest observed resolution are excluded from spawning and receive reduced update weights.
Finally, a targeted repair pass allocates 1000 iterations to the most deteriorated training views, identified by photometric errors exceeding (1.5$\times$) the median, mitigating recency bias.

\begin{figure*}[]
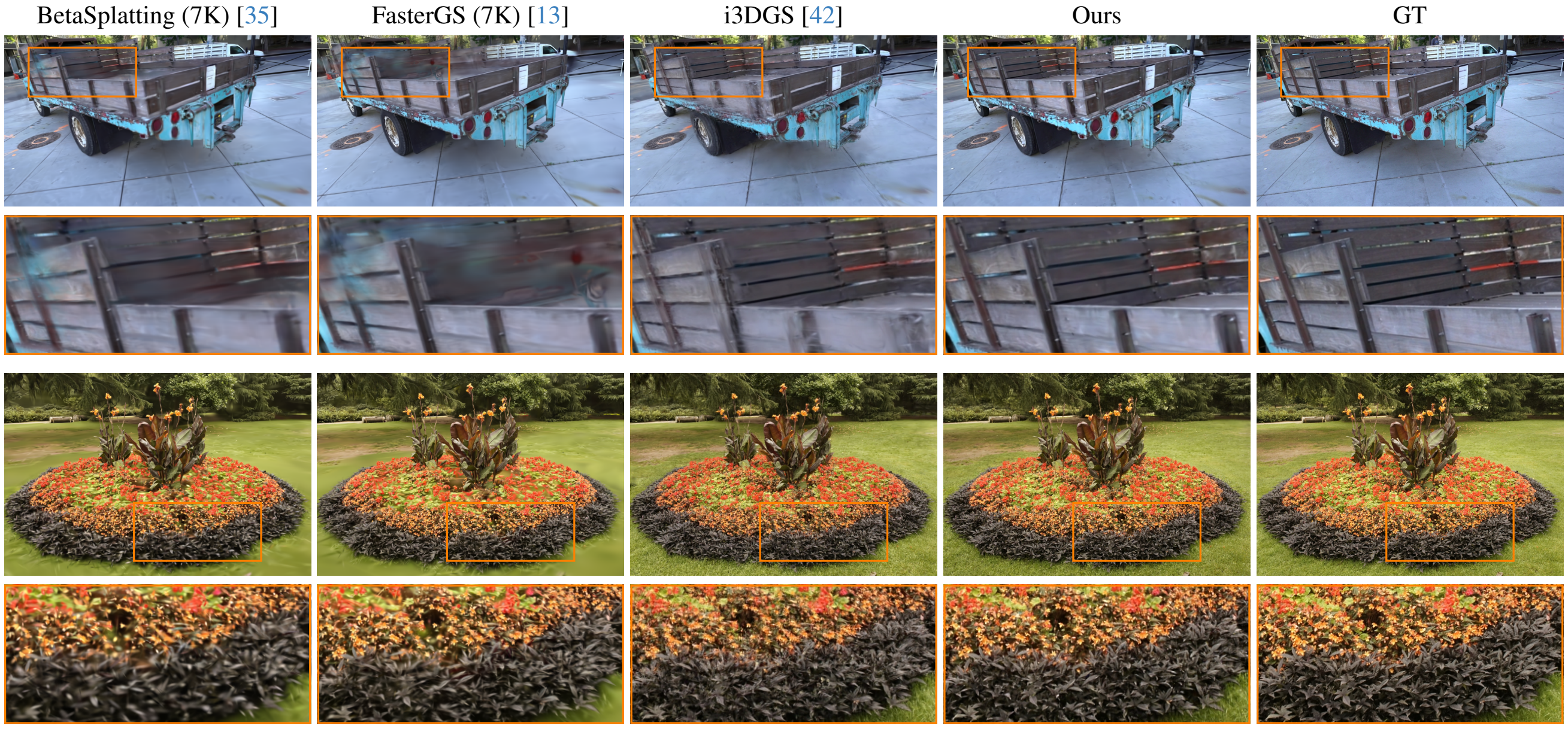

  \centering%
  \setlength{\tabcolsep}{1pt}%
  \setlength{\cmpw}{0.194\textwidth}%
  \footnotesize%
  \vspace{-5mm}
  \begin{tabular}{ccccc}%
    BetaSplatting (7K)~\cite{liu2025dbs} & FasterGS (7K)~\cite{hahlbohm2026fastergs} & i3DGS~\cite{meuleman2026immediate} & Ours & GT \\[1pt]\cmpscenefive{mip_tandt_db/tandt_truck} \cmpscenefive{mip_tandt_db/MipNeRF360U_bonus_flowers}\end{tabular} \vspace{-5mm} \caption{Qualitative comparison against baselines on the 3DGS benchmark: \textsc{Truck}, and \textsc{Flowers} scenes.
  }%
  \label{fig:comp_offline}%
\end{figure*}
\providecommand{\rotrow}[1]{%
  \multirow{4}{25pt}{\rotatebox[origin=c]{90}{{#1}}}}
\begin{table*}[!h]
  \setlength{\tabcolsep}{3pt}
  \caption{\label{tab:gs_benchmark}
    \textbf{Results on the full 3DGS benchmark}.
    We compare with online method i3DGS as well as FasterGS and BetaSplatting with GLOMAP sequential mapping (requiring 1:52 on MipNeRF360, 1:42 on Tanks\&Temples and failing on Deep Blending).
    More expensive exhaustive mapping is in \suppref{tab:gs_benchmark_full}.
    The {\colorbox{firstcolor}{best}} and {\colorbox{secondcolor}{second best}} are color coded.
  }
  \vspace{-7pt}
  \setlength{\tabcolsep}{3pt}
  \footnotesize

  \centering
  \begin{tabular}{@{}c@{\hspace{-12pt}}l|cccc|cccc|cccc}\toprule
                                                    &                     &
    \multicolumn{4}{c|}{\textsc{MipNeRF360}}        &
    \multicolumn{4}{c|}{\textsc{Tanks and Temples}} &
    \multicolumn{4}{c}{\textsc{Deep Blending}}                                                                                                                                                                                                                                                          \\
    {}                                    &                     & PSNR$^\uparrow$    & SSIM$^\uparrow$   & LPIPS$^\downarrow$ & Time$^\downarrow$ &
    PSNR$^\uparrow$                                 & SSIM$^\uparrow$     & LPIPS$^\downarrow$ & Time$^\downarrow$ &
    PSNR$^\uparrow$                                 & SSIM$^\uparrow$     & LPIPS$^\downarrow$ & Time$^\downarrow$                                                                                                                                                                                      \\\hline

    \rotrow{G (\textit{seq.)}}                      & BetaSplatting (30K) & \first{}25.90      & 0.703             & 0.333              & 0:09:07           & \first{}23.31  & \first{}0.831  & \first{}0.190  & 0:08:27          & $\times$       & $\times$       & $\times$       & $\times$         \\
                                                    & FasterGS (30K)      & \second{}25.53     & 0.702             & 0.356              & 0:05:29           & 22.17          & 0.802          & 0.250          & 0:04:21          & $\times$       & $\times$       & $\times$       & $\times$         \\
                                                    & BetaSplatting (7K)  & 24.05              & 0.656             & 0.396              & 0:03:17           & 20.81          & 0.733          & 0.321          & 0:02:55          & $\times$       & $\times$       & $\times$       & $\times$         \\
                                                    & FasterGS (7K)       & 24.57              & 0.678             & 0.411              & 0:02:29           & 20.79          & 0.742          & 0.343          & 0:02:10          & $\times$       & $\times$       & $\times$       & $\times$         \\

    \hline
    \multirow{2}{20pt}                              & i3DGS               & 24.75              & \second{}0.721    & \second{}0.280     & \second{}0:01:59  & 22.08          & 0.783          & 0.226          & \second{}0:01:25 & \first{}27.16  & \first{}0.856  & \first{}0.285  & \second{}0:02:03 \\
    \cline{2-14}
                                                    & Ours                & 25.19              & \first{}0.731     & \first{}0.264      & \first{}0:01:17   & \second{}23.00 & \second{}0.803 & \second{}0.200 & \first{}0:01:18  & \second{}26.49 & \second{}0.847 & \second{}0.295 & \first{}0:01:22  \\
    \bottomrule
  \end{tabular}
\end{table*}

\section{Results and Experiments}
\label{sec:experiments}

We evaluate using the online protocols of Meuleman \etal.~\cite{meuleman2025onthefly} on three MipNeRF360 scenes (garden, counter, bonsai, marked with \textit{otf})~\cite{barron2022mipnerf360}, three TUM scenes~\cite{sturm2012benchmark}, and three StaticHikes captures~\cite{meuleman2023progressively}, with the same held-out views for evaluation.
We additionally evaluate on the 3DGS benchmark~\cite{kerbl3dgs,barron2022mipnerf360,hedman2018deepblending,knapitsch2017tanks} and two large-scale captures, \textsc{SmallCity}~\cite{kerbl2024hierarchical} and \textsc{CityWalk}~\cite{meuleman2025onthefly}.
We compare to methods including MonoGS~\cite{matsuki2024monogs}, S3PO-GS~\cite{cheng2025s3pogs}, LongSplat~\cite{lin2025longsplat}, On-The-Fly NVS~\cite{meuleman2025onthefly}, i3DGS~\cite{meuleman2026immediate}, DROID-Splat~\cite{homeyer2024droid}, CF-3DGS~\cite{Fu_2024_CVPR}, DepthAnything3~\cite{lin2025depthanything3}, ARTDECO~\cite{li2026artdeco}, H3DGS~\cite{kerbl2024hierarchical} and Octree-GS~\cite{ren2024octree}, as well as GLOMAP-based~\cite{pan2024glomap,schoenberger2016sfm} offline references using Beta Splatting~\cite{liu2025dbs} and Faster-GS~\cite{hahlbohm2026fastergs}.
We report PSNR, SSIM, LPIPS, and total time, averaging ours over five runs.

Unless noted otherwise, experiments use an RTX 5090 (32\,GB) and i9-14900KS with 64GB DDR4 RAM on Windows 11.
For methods requiring reduced input resolution, we follow~\cite{meuleman2025onthefly,meuleman2026immediate}.
These results are reported in \suppref{sec:suppl-additional-res}.

\subsection{Results}
\label{sec:exp-main}

\cref{tab:slam-eval-s3po} evaluates on the MipNeRF360 (\textit{otf}), TUM, and StaticHikes scenes against pose-free and online baselines.
In this benchmark, our method outperforms our closest competitor i3DGS (except time on TUM), and is consistently among the strongest methods overall.
On MipNeRF360 (\textit{otf}) and StaticHikes, we surpass competing pose-free methods on all three metrics while reducing computation time by up to $40\%$.
On TUM, LongSplat has slightly better PSNR and SSIM, but takes roughly $50\times$ longer.

On the standard 3DGS benchmark (\cref{tab:gs_benchmark,fig:comp_offline}), we compare against BetaSplatting and FasterGS with sequential (fast) GLOMAP pose estimation (more expensive exhaustive setting in \suppref{tab:gs_benchmark_full}).
Our method is consistently the fastest, outperforming in SSIM and LPIPS on MipNeRF360 while remaining competitive on Tanks\&Temples.
DeepBlending is more challenging; however replacing spherical-beta colors with SH improves our result by $\sim0.3$\,dB (\suppref{sec:suppl-shdb}).
On MipNeRF360, sequential matching can drop frames (e.g., bicycle), and metrics are therefore computed only on successfully registered views.
Against exhaustive-GLOMAP references (\suppref{tab:gs_benchmark_full}), we remain competitive at 7k iterations, while 30k iterations achieve higher quality at $\sim6$--$12\times$ our runtime and require the complete capture to be available beforehand.

\begin{table}[]
  \caption{\label{tab:large-scale}
    \textbf{Results for large scale scenes.}
    Offline methods (H3DGS, Octree-GS) include COLMAP times.
    Comparing image metrics are from~\cite{meuleman2026immediate}, non-approximate times measured on our machine.
    S3PO-GS includes refinement.
    The {\colorbox{firstcolor}{best}} and {\colorbox{secondcolor}{second best}} are color coded.
  }
  \vspace{-7pt}
  \setlength{\tabcolsep}{1pt}
  \footnotesize
  \begin{tabular}{l|cccr|cccr}
    \toprule
                                            &
    \multicolumn{4}{c|}{\textsc{SmallCity}} &
    \multicolumn{4}{c}{\textsc{CityWalk}}                                                                                                                                                     \\
                                            &
    PSNR$^\uparrow$                         & SSIM$^\uparrow$  & LPIPS$^\downarrow$ & Time$^\downarrow$ &
    PSNR$^\uparrow$                         & SSIM$^\uparrow$  & LPIPS$^\downarrow$ & Time$^\downarrow$                                                                                       \\
    \midrule
    {H3DGS}                                &
    21.17                                   & 0.679            & \second{}0.285     & $\sim$2.5h        & 11.78          & 0.557          & 0.560          & $\sim$20h                        \\
    Octree-GS                               & \second{}{24.18} & \first{}{0.831}    & \first{}{0.273}   & $\sim$1h       & {17.33}        & {0.638}        & {0.520}         & $\sim$2.5h     \\
    \hline\hline
    S3PO-GS                                 & {19.71}          & {0.652}            & {0.462}           & $\sim$0.7h     & {10.72}        & {0.439}        & {0.579}         & $\sim$5.5h     \\
    {OtF NVS}                               & 23.59            & 0.789              & 0.323             & \first{}02:21  & 21.71          & 0.712          & 0.395           & \first{}29:35  \\
    i3DGS                                   & 23.56            & 0.800              & 0.302             & \second{}02:54 & \second{}22.03 & \second{}0.772 & \second{} 0.336 & 45:20          \\\hline
    Ours                                    & \first{}24.45    & \second{}0.808     & \second{}0.285    & 03:47          & \first{}25.81  & \first{}0.857  & \first{} 0.239  & \second{}42:43 \\
    \bottomrule
  \end{tabular}
\end{table}

\cref{tab:large-scale} evaluates the large scale SmallCity and CityWalk captures.
Our method achieves the best metric on both scenes except for SSIM and LPIPS on SmallCity, where Octree-GS is marginally better but requires about one hour versus roughly 4 minutes for us.
Compared with On-The-Fly NVS, we are $\sim50\%$ slower while improving PSNR by $0.86$\,dB on SmallCity and $4.1$\,dB on CityWalk.

\subsection{Ablation Studies} \label{sec:ablation} \begin{table}[t]
  \centering\footnotesize\setlength{\tabcolsep}{3pt}
  \caption{\textbf{Component ablation} (leave-one-out from full method).
    Top table: Metrics are averaged over all nine OtF scenes (\cref{tab:slam-eval-s3po}).
  }
  \label{tab:abl-flagship}
  \begin{tabular}{l c|c c c r}
    \toprule
    Configuration           &                          & PSNR$\uparrow$ & SSIM$\uparrow$ & LPIPS$\downarrow$ & t\,(s) \\
    \midrule
    \textbf{Full (ours)}    &                          & \textbf{25.30} & \textbf{0.810} & \textbf{0.213}    & {101}  \\
    \midrule
    w/o SB lobes ($\to$ SH) & [\cref{sec:beta_prim}]   & 25.12          & 0.807          & 0.214             & 106    \\
    DBS lobe init prior     & [\cref{sec:beta_prim}]   & 25.19          & 0.809          & 0.214             & 100    \\
    w/o confidence          & [\cref{sec:spawn}]       & 25.14          & 0.808          & 0.215             & 99     \\
    w/o solidification      & [\cref{sec:solidity}]    & 25.15          & 0.807          & 0.216             & 100    \\
    w/o error relocation    & [\cref{sec:error_reloc}] & 25.09          & 0.804          & 0.219             & 95     \\
    w/o regularizers        & [\cref{sec:optim}]       & 24.98          & 0.803          & 0.216             & 108    \\
    w/o fused rasterizer    & [\cref{sec:raster}]      & 25.24          & 0.809          & 0.214             & 120    \\
    \bottomrule
  \end{tabular}
  \centering\footnotesize\setlength{\tabcolsep}{1.02pt}
  \begin{tabular}{l|ccc|ccc}
    \toprule
    Large-scene confs                       &
    \multicolumn{3}{c|}{\textsc{SmallCity}} &
    \multicolumn{3}{c}{\textsc{CityWalk}}                                                                                                              \\
    \text{[}\cref{sec:hierarchy}\text{]}    &
    PSNR$^\uparrow$                         & SSIM$^\uparrow$ & LPIPS$^\downarrow$ &
    PSNR$^\uparrow$                         & SSIM$^\uparrow$ & LPIPS$^\downarrow$                                                                     \\
    \midrule
    \textbf{Full (ours)}                    & \textbf{24.45}  & \textbf{0.808}     & \textbf{0.285} & \textbf{25.81} & \textbf{0.857} & \textbf{0.239} \\
    \midrule
    w/o moment merge                        & 23.78           & 0.800              & 0.299          & 23.54          & 0.833          & 0.262          \\
    w/o error repair                        & 23.60           & 0.798              & 0.295          & 23.44          & 0.843          & 0.244          \\
    w/o res-weighting                       & 24.12           & 0.805              & 0.294          & 22.93          & 0.812          & 0.294          \\
    \bottomrule
  \end{tabular}
\end{table}
\begin{table}[t]
  \centering\footnotesize\setlength{\tabcolsep}{2pt}
  \caption{\textbf{Kernel ablation}.
    Metrics averaged over all OtF scenes.
  }
  \label{tab:abl-kernel}
  \begin{tabular}{l c|c c c r}
    \toprule
    Configuration                      & Confidence                                                      & PSNR$\uparrow$ & SSIM$\uparrow$ & LPIPS$\downarrow$ & t\,(s) \\
    \midrule
    i3DGS~\cite{meuleman2026immediate} & none                                                            & 24.73          & 0.797          & 0.227             & 129    \\
    Gaussian                           & $o_{\mathrm{spawn}} $                                           & 25.09          & 0.808          & 0.214             & 130    \\
    $\beta$ frozen at 4                & $o_{\mathrm{spawn}}$                                            & 25.06          & 0.806          & 0.217             & 100    \\
    learnable $\beta$                  & $o_{\mathrm{spawn}} $                                           & 25.11          & 0.807          & 0.215             & 100    \\
    learnable $\beta$                  & $o_{\mathrm{spawn}}$,  $\beta_{\mathrm{spawn}}$                 & 25.15          & 0.807          & 0.216             & 100    \\
    \textbf{Full (ours)}               & $o_{\mathrm{spawn}}$,  $\beta_{\mathrm{spawn}}$,\text{solidify} & \textbf{25.30} & \textbf{0.810} & \textbf{0.213}    & {101}  \\
    \bottomrule
  \end{tabular}
\end{table}

\cref{tab:abl-flagship} (top) ablates the components of our final configuration across the nine \textit{otf} scenes.
Spherical-beta lobes (SB) provide a small quality and runtime improvement over SH, although SH is preferable on scenes such as Deep Blending (\suppref{sec:suppl-shdb}).
Confidence-aware spawning, solidification, and zero-initialized lobes each improve reconstruction quality.
The fused rasterizer reduces runtime by about $16\%$; the 30-iteration optimization burst reduces from 287ms to 170ms (\textsc{Garden}).
Error-driven relocation yields a small runtime penalty of $\sim6\%$ but improves quality.
The opacity and scale regularizers further improve both runtime and quality, by feeding ineffective primitives into relocation.

\cref{tab:abl-flagship} (bottom) evaluates the large-scale hierarchical setting (\cref{sec:hierarchy}).
Our moment-based merge consistently outperforms the opacity-weighted baseline by avoiding overly large parent primitives, with the largest gain on \textsc{CityWalk} ($+2.27$\,dB) where multiple hierarchy levels are required.
Resolution-aware weighting and error-based repair further improve quality by preventing late, low-resolution observations from degrading previously well-reconstructed regions.

\cref{tab:abl-kernel} evaluates the effect of different kernel choices on the nine \textit{otf} scenes.
Beta kernels provide a small quality improvement over Gaussians, while also being faster due to their bounded support.
Together with confidence-aware spawning, they provide a $+0.2$\,dB improvement.

\subsection{Quality During Reconstruction}
\label{sec:exp-online}
\begin{figure}
  \centering
  \IfFileExists{figures/online_study/trajectory.pdf}{%
    \includegraphics[width=\linewidth]{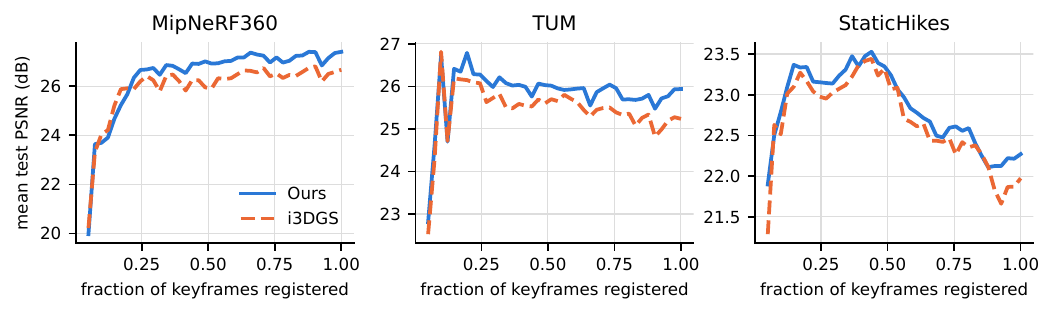}}{%
    \fbox{\parbox{0.95\linewidth}{\centering\vspace{1.5em}\outofdate trajectory.pdf (running)\vspace{1.5em}}}}\\[2pt]
  \IfFileExists{figures/online_study/immediate_vs_settled2.pdf}{%
    \includegraphics[width=\linewidth]{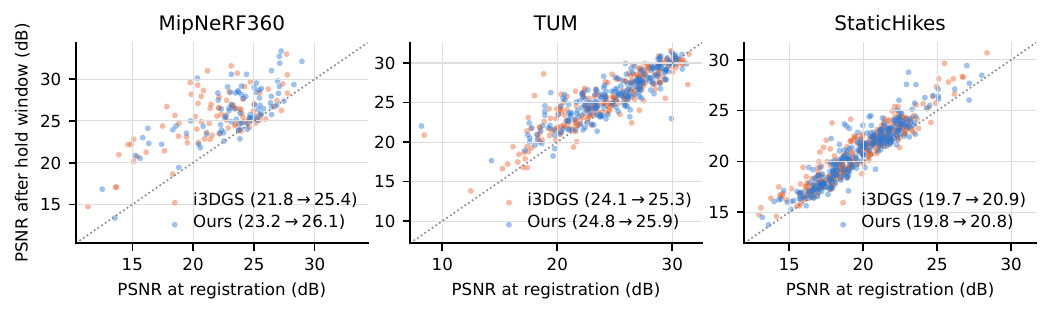}}{%
    \fbox{\parbox{0.95\linewidth}{\centering\vspace{1.5em}\outofdate immediate\_vs\_settled.pdf (running)\vspace{1.5em}}}}
  \vspace{-7mm}
  \caption{\textbf{Quality during reconstruction.}
    \emph{Top}: mean PSNR over test views registered so far versus reconstruction progress, for our method and i3DGS.
    \emph{Bottom}: PSNR immediately after each view's registration versus just before the next test view arrives.
    Points above the diagonal show improvement after registration.
  }
  \label{fig:online-quality}
\end{figure}

An online system should improve throughout reconstruction, not only in the final model.
\cref{fig:online-quality} (top) reports mean PSNR over the test views registered so far as reconstruction progresses.
Our method consistently outperforms i3DGS on MipNeRF360 \textit{(otf)}, TUM, and StaticHikes, with the gap emerging shortly after bootstrap and persisting throughout the capture.
The bottom row measures how much subsequent keyframes improve a \textit{test} view after it enters the map: each point compares its PSNR immediately after registration with its PSNR just before the next held-out view is registered.
Nearly all points lie above the diagonal, indicating continued improvement without forgetting.

\section{Conclusion}
\label{sec:conclusion}\label{sec:limitations}

\paragraph{Limitations and Future Work.}
On the 3DGS benchmark, experiments show a remaining quality gap to fully optimized offline methods, suggesting that further improvements in online pose refinement~\cite{meuleman2026immediate}---orthogonal to our work---are a promising direction for closing the remaining gap.
The merge approximates out-of-band groups and the rendered contribution, and shuffled input can reduce quality.
Further promising directions include integration with VR~\cite{franke2025vrsplatting} and using the confidence map for refinement~\cite{goli2024bayesrays}.

\paragraph{Conclusion.}
We present an online reconstruction framework that redesigns the primitive lifecycle around the short optimization horizon of incremental capture.
Our bounded beta kernel encodes a continuous notion of solidity, observability-aware spawning initializes this confidence from geometric evidence, and error-driven relocation reallocates unused capacity to poorly explained regions.
Closed-form moment matching enables a shape-aware hierarchy, while bounded-support rasterization and fused optimization make the representation efficient for large-scale online reconstruction.
Together, these components yield faster and higher-quality reconstruction throughout capture, while turning the primitive shape field into a live confidence visualization.

\section*{Acknowledgments}
We thank the anonymous reviewers for their constructive feedback.

{
	\small
	\bibliographystyle{ieeenat_fullname}
	\bibliography{references}

@inproceedings{meuleman2026immediate,
  author    = {Meuleman, Andreas and Franke, Linus and Zhestiankin, Boris and Montemagni, Camille and Drettakis, George},
  title     = {Immediate {3D} {G}aussian Splat Reconstruction of Unordered Input with Global Consistency},
  booktitle = {ACM SIGGRAPH 2026 Conference Papers},
  year      = {2026},
  doi       = {10.1145/3799902.3811167}
}

@article{meuleman2025onthefly,
  author    = {Meuleman, Andreas and Shah, Ishaan and Lanvin, Alexandre and Kerbl, Bernhard and Drettakis, George},
  title     = {On-the-fly Reconstruction for Large-Scale Novel View Synthesis from Unposed Images},
  journal   = {ACM Transactions on Graphics},
  volume    = {44},
  number    = {4},
  year      = {2025},
  doi       = {10.1145/3730913}
}

@article{kerbl3dgs,
  author    = {Kerbl, Bernhard and Kopanas, Georgios and Leimk{\"u}hler, Thomas and Drettakis, George},
  title     = {{3D} {G}aussian Splatting for Real-Time Radiance Field Rendering},
  journal   = {ACM Transactions on Graphics},
  volume    = {42},
  number    = {4},
  year      = {2023},
  doi       = {10.1145/3592433}
}

@inproceedings{mildenhall2020nerf,
  author    = {Mildenhall, Ben and Srinivasan, Pratul P. and Tancik, Matthew and Barron, Jonathan T. and Ramamoorthi, Ravi and Ng, Ren},
  title     = {{NeRF}: Representing Scenes as Neural Radiance Fields for View Synthesis},
  booktitle = {European Conference on Computer Vision (ECCV)},
  year      = {2020}
}

@inproceedings{barron2022mipnerf360,
  author    = {Barron, Jonathan T. and Mildenhall, Ben and Verbin, Dor and Srinivasan, Pratul P. and Hedman, Peter},
  title     = {Mip-{NeRF} 360: Unbounded Anti-Aliased Neural Radiance Fields},
  booktitle = {IEEE/CVF Conference on Computer Vision and Pattern Recognition (CVPR)},
  year      = {2022}
}

@article{kerbl2024hierarchical,
  author    = {Kerbl, Bernhard and Meuleman, Andreas and Kopanas, Georgios and Wimmer, Michael and Lanvin, Alexandre and Drettakis, George},
  title     = {A Hierarchical {3D} {G}aussian Representation for Real-Time Rendering of Very Large Datasets},
  journal   = {ACM Transactions on Graphics},
  volume    = {43},
  number    = {4},
  year      = {2024}
}

@inproceedings{mallick2024taming,
  author    = {Mallick, Saswat Subhajyoti and Goel, Rahul and Kerbl, Bernhard and Steinberger, Markus and Carrasco, Francisco Vicente and De La Torre, Fernando},
  title     = {Taming {3DGS}: High-Quality Radiance Fields with Limited Resources},
  booktitle = {SIGGRAPH Asia 2024 Conference Papers},
  year      = {2024},
  doi       = {10.1145/3680528.3687694}
}

@article{durvasula2024distwar,
  author    = {Durvasula, Sankeerth and Zhao, Adrian and Chen, Fan and Liang, Ruofan and Sanjaya, Pawan Kumar and Vijaykumar, Nandita},
  title     = {{DISTWAR}: Fast Differentiable Rendering on Raster-based Rendering Pipelines},
  journal   = {arXiv preprint arXiv:2401.05345},
  year      = {2024}
}

@inproceedings{hahlbohm2026fastergs,
  author    = {Hahlbohm, Florian and Franke, Linus and Eisemann, Martin and Magnor, Marcus},
  title     = {Faster-{GS}: Analyzing and Improving {G}aussian Splatting Optimization},
  booktitle = {IEEE/CVF Conference on Computer Vision and Pattern Recognition (CVPR)},
  year      = {2026}
}

@inproceedings{hanson2025speedy,
  author    = {Hanson, Alex and Tu, Allen and Lin, Geng and Singla, Vasu and Zwicker, Matthias and Goldstein, Tom},
  title     = {{Speedy-Splat}: Fast {3D} {G}aussian Splatting with Sparse Pixels and Sparse Primitives},
  booktitle = {IEEE/CVF Conference on Computer Vision and Pattern Recognition (CVPR)},
  year      = {2025}
}

@inproceedings{kheradmand2024mcmc,
  author    = {Kheradmand, Shakiba and Rebain, Daniel and Sharma, Gopal and Sun, Weiwei and Tseng, Yang-Che and Isack, Hossam and Kar, Abhishek and Tagliasacchi, Andrea and Yi, Kwang Moo},
  title     = {{3D} {G}aussian Splatting as {M}arkov {C}hain {M}onte {C}arlo},
  booktitle = {Advances in Neural Information Processing Systems (NeurIPS)},
  year      = {2024}
}

@inproceedings{rotabulo2024revising,
  author    = {Rota Bul{\`o}, Samuel and Porzi, Lorenzo and Kontschieder, Peter},
  title     = {Revising Densification in {G}aussian Splatting},
  booktitle = {European Conference on Computer Vision (ECCV)},
  year      = {2024}
}

@inproceedings{fang2024minisplatting,
  author    = {Fang, Guangchi and Wang, Bing},
  title     = {{Mini-Splatting}: Representing Scenes with a Constrained Number of {G}aussians},
  booktitle = {European Conference on Computer Vision (ECCV)},
  year      = {2024}
}

@inproceedings{liu2025dbs,
  author    = {Liu, Rong and Sun, Dylan and Chen, Meida and Wang, Yue and Feng, Andrew},
  title     = {Deformable {B}eta Splatting},
  booktitle = {ACM SIGGRAPH 2025 Conference Papers},
  year      = {2025},
  doi       = {10.1145/3721238.3730716}
}

@inproceedings{hamdi2024ges,
  author    = {Hamdi, Abdullah and Melas-Kyriazi, Luke and Mai, Jinjie and Qian, Guocheng and Liu, Ruoshi and Vondrick, Carl and Ghanem, Bernard and Vedaldi, Andrea},
  title     = {{GES}: Generalized Exponential Splatting for Efficient Radiance Field Rendering},
  booktitle = {IEEE/CVF Conference on Computer Vision and Pattern Recognition (CVPR)},
  year      = {2024}
}

@inproceedings{huang20242dgs,
  author    = {Huang, Binbin and Yu, Zehao and Chen, Anpei and Geiger, Andreas and Gao, Shenghua},
  title     = {{2D} {G}aussian Splatting for Geometrically Accurate Radiance Fields},
  booktitle = {ACM SIGGRAPH 2024 Conference Papers},
  year      = {2024}
}

@inproceedings{vonluetzow2025linprim,
  author    = {von L{\"u}tzow, Nicolas and Nie{\ss}ner, Matthias},
  title     = {{LinPrim}: Linear Primitives for Differentiable Volumetric Rendering},
  booktitle = {Advances in Neural Information Processing Systems (NeurIPS)},
  year      = {2025}
}

@inproceedings{mai2025ever,
  author    = {Mai, Alexander and Hedman, Peter and Kopanas, Georgios and Verbin, Dor and Futschik, David and Xu, Qiangeng and Kuester, Falko and Barron, Jonathan T. and Zhang, Yinda},
  title     = {{EVER}: Exact Volumetric Ellipsoid Rendering for Real-time View Synthesis},
  booktitle = {IEEE/CVF International Conference on Computer Vision (ICCV)},
  year      = {2025}
}

@inproceedings{held2025convex,
  author    = {Held, Jan and Vandeghen, Renaud and Hamdi, Abdullah and Deliege, Adrien and Cioppa, Anthony and Giancola, Silvio and Vedaldi, Andrea and Ghanem, Bernard and Van Droogenbroeck, Marc},
  title     = {{3D} Convex Splatting: Radiance Field Rendering with {3D} Smooth Convexes},
  booktitle = {IEEE/CVF Conference on Computer Vision and Pattern Recognition (CVPR)},
  year      = {2025}
}

@inproceedings{li2025streamgs,
  author    = {Li, Yang and Wang, Jinglu and Chu, Lei and Li, Xiao and Kao, Shiu-hong and Chen, Ying-Cong and Lu, Yan},
  title     = {{StreamGS}: Online Generalizable {G}aussian Splatting Reconstruction for Unposed Image Streams},
  booktitle = {IEEE/CVF International Conference on Computer Vision (ICCV)},
  year      = {2025}
}

@inproceedings{wang2025vggt,
  author    = {Wang, Jianyuan and Chen, Minghao and Karaev, Nikita and Vedaldi, Andrea and Rupprecht, Christian and Novotny, David},
  title     = {{VGGT}: Visual Geometry Grounded Transformer},
  booktitle = {IEEE/CVF Conference on Computer Vision and Pattern Recognition (CVPR)},
  year      = {2025}
}

@inproceedings{matsuki2024monogs,
  author    = {Matsuki, Hidenobu and Murai, Riku and Kelly, Paul H. J. and Davison, Andrew J.},
  title     = {{G}aussian Splatting {SLAM}},
  booktitle = {IEEE/CVF Conference on Computer Vision and Pattern Recognition (CVPR)},
  year      = {2024}
}

@inproceedings{keetha2024splatam,
  author    = {Keetha, Nikhil and Karhade, Jay and Jatavallabhula, Krishna Murthy and Yang, Gengshan and Scherer, Sebastian and Ramanan, Deva and Luiten, Jonathon},
  title     = {{SplaTAM}: Splat, Track \& Map {3D} {G}aussians for Dense {RGB-D} {SLAM}},
  booktitle = {IEEE/CVF Conference on Computer Vision and Pattern Recognition (CVPR)},
  year      = {2024}
}

@inproceedings{huang2024photoslam,
  author    = {Huang, Huajian and Li, Longwei and Cheng, Hui and Yeung, Sai-Kit},
  title     = {Photo-{SLAM}: Real-time Simultaneous Localization and Photorealistic Mapping for Monocular, Stereo, and {RGB-D} Cameras},
  booktitle = {IEEE/CVF Conference on Computer Vision and Pattern Recognition (CVPR)},
  year      = {2024}
}

@inproceedings{peng2024rtgslam,
  author    = {Peng, Zhexi and Shao, Tianjia and Liu, Yong and Zhou, Jingke and Yang, Yin and Wang, Jingdong and Zhou, Kun},
  title     = {{RTG-SLAM}: Real-time {3D} Reconstruction at Scale using {G}aussian Splatting},
  booktitle = {ACM SIGGRAPH 2024 Conference Papers},
  year      = {2024},
  doi       = {10.1145/3641519.3657455}
}

@inproceedings{zhu2025loopsplat,
  author    = {Zhu, Liyuan and Li, Yue and Sandstr{\"o}m, Erik and Huang, Shengyu and Schindler, Konrad and Armeni, Iro},
  title     = {{LoopSplat}: Loop Closure by Registering {3D} {G}aussian Splats},
  booktitle = {International Conference on 3D Vision (3DV)},
  year      = {2025}
}

@inproceedings{schoenberger2016sfm,
  author    = {Sch{\"o}nberger, Johannes L. and Frahm, Jan-Michael},
  title     = {Structure-from-Motion Revisited},
  booktitle = {IEEE/CVF Conference on Computer Vision and Pattern Recognition (CVPR)},
  year      = {2016}
}

@inproceedings{alibey2023mixvpr,
  author    = {Ali-bey, Amar and Chaib-draa, Brahim and Gigu{\`e}re, Philippe},
  title     = {{MixVPR}: Feature Mixing for Visual Place Recognition},
  booktitle = {IEEE/CVF Winter Conference on Applications of Computer Vision (WACV)},
  year      = {2023}
}

@inproceedings{potje2024xfeat,
  author    = {Potje, Guilherme and Cadar, Felipe and Araujo, Andr{\'e} and Martins, Renato and Nascimento, Erickson R.},
  title     = {{XFeat}: Accelerated Features for Lightweight Image Matching},
  booktitle = {IEEE/CVF Conference on Computer Vision and Pattern Recognition (CVPR)},
  year      = {2024}
}

@inproceedings{lindenberger2023lightglue,
  author    = {Lindenberger, Philipp and Sarlin, Paul-Edouard and Pollefeys, Marc},
  title     = {{LightGlue}: Local Feature Matching at Light Speed},
  booktitle = {IEEE/CVF International Conference on Computer Vision (ICCV)},
  year      = {2023}
}

@inproceedings{yang2024depthanythingv2,
  author    = {Yang, Lihe and Kang, Bingyi and Huang, Zilong and Zhao, Zhen and Xu, Xiaogang and Feng, Jiashi and Zhao, Hengshuang},
  title     = {{Depth Anything V2}},
  booktitle = {Advances in Neural Information Processing Systems (NeurIPS)},
  year      = {2024}
}

@inproceedings{lin2025depthanything3,
 author = {Lin, Haotong and Chen, Sili and Liew, Jun Hao and Chen, Donny Y. and Li, Zhenyu and Zhao, Yang and Peng, Sida and Guo, Hengkai and Zhou, Xiaowei and Shi, Guang and Feng, Jiashi and Kang, Bingyi},
 booktitle = {International Conference on Learning Representations (ICLR)},
 title = {{Depth Anything 3}: Recovering the Visual Space from Any Views},
 url = {https://proceedings.iclr.cc/paper_files/paper/2026/file/e4cd50120b6d7e8daff1749d6bbaa889-Paper-Conference.pdf},
 year = {2026}
}

@inproceedings{goli2024bayesrays,
  author    = {Goli, Lily and Reading, Cody and Sell{\'a}n, Silvia and Jacobson, Alec and Tagliasacchi, Andrea},
  title     = {{B}ayes' Rays: Uncertainty Quantification for Neural Radiance Fields},
  booktitle = {IEEE/CVF Conference on Computer Vision and Pattern Recognition (CVPR)},
  year      = {2024}
}

@inproceedings{jiang2024fisherrf,
  author    = {Jiang, Wen and Lei, Boshu and Daniilidis, Kostas},
  title     = {{FisherRF}: Active View Selection and Uncertainty Quantification for Radiance Fields using {F}isher Information},
  booktitle = {European Conference on Computer Vision (ECCV)},
  year      = {2024}
}

@article{knapitsch2017tanks,
  author    = {Knapitsch, Arno and Park, Jaesik and Zhou, Qian-Yi and Koltun, Vladlen},
  title     = {{Tanks and Temples}: Benchmarking Large-Scale Scene Reconstruction},
  journal   = {ACM Transactions on Graphics},
  volume    = {36},
  number    = {4},
  year      = {2017}
}

@article{hedman2018deepblending,
  author  = {Hedman, Peter and Philip, Julien and Price, True and Frahm, Jan-Michael and Drettakis, George and Brostow, Gabriel},
  title   = {Deep Blending for Free-Viewpoint Image-Based Rendering},
  journal = {ACM Transactions on Graphics},
  volume  = {37},
  number  = {6},
  year    = {2018}
}

@inproceedings{lin2025longsplat,
  author    = {Lin, Chin-Yang and Sun, Cheng and Yang, Fu-En and Chen, Min-Hung and Lin, Yen-Yu and Liu, Yu-Lun},
  title     = {{LongSplat}: Robust Unposed {3D} {G}aussian Splatting for Casual Long Videos},
  booktitle = {IEEE/CVF International Conference on Computer Vision (ICCV)},
  year      = {2025}
}

@article{jiang2025anysplat,
  title={{AnySplat}: Feed-forward {3D} {G}aussian Splatting from Unconstrained Views},
  author={Jiang, Lihan and Mao, Yucheng and Xu, Linning and Lu, Tao and Ren, Kerui and Jin, Yichen and Xu, Xudong and Yu, Mulin and Pang, Jiangmiao and Zhao, Feng and others},
  journal={ACM Transactions on Graphics},
  volume={44},
  number={6},
  year={2025}
}

@inproceedings{cheng2025s3pogs,
  author    = {Cheng, Chong and Yu, Sicheng and Wang, Zijian and Zhou, Yifan and Wang, Hao},
  title     = {Outdoor Monocular {SLAM} with Global Scale-Consistent {3D} {G}aussian Pointmaps},
  booktitle = {IEEE/CVF International Conference on Computer Vision (ICCV)},
  year      = {2025}
}

@inproceedings{li2026artdeco,
 author = {Li, Guanghao and Ren, Kerui and Xu, Linning and Zheng, Zhewen and Jiang, Changjian and Gao, Xin and DAI, Bo and Pu, Jian and Yu, Mulin and Pang, Jiangmiao},
 booktitle = {International Conference on Learning Representations},
 pages = {102997--103029},
 title = {ARTDECO: Toward High-Fidelity On-the-Fly Reconstruction with Hierarchical Gaussian Structure and Feed-Forward Guidance},
 volume = {2026},
 year = {2026}
}

@article{ren2024octree,
  title={{Octree-GS}: Towards Consistent Real-Time Rendering with {LOD}-Structured {3D} {G}aussians},
  author={Ren, Kerui and Jiang, Lihan and Lu, Tao and Yu, Mulin and Xu, Linning and Ni, Zhangkai and Dai, Bo},
  journal={IEEE Transactions on Pattern Analysis and Machine Intelligence},
  year={2025}
}

@article{homeyer2024droid,
  title={{DROID-Splat}: Combining End-to-End {SLAM} with {3D} {G}aussian Splatting},
  author={Homeyer, Christian and Begiristain, Leon and Schn{\"o}rr, Christoph},
  journal={arXiv preprint arXiv:2411.17660},
  year={2024}
}

@inproceedings{Fu_2024_CVPR,
  author    = {Fu, Yang and Liu, Sifei and Kulkarni, Amey and Kautz, Jan and Efros, Alexei A. and Wang, Xiaolong},
  title     = {{COLMAP}-Free {3D} {G}aussian Splatting},
  booktitle = {IEEE/CVF Conference on Computer Vision and Pattern Recognition (CVPR)},
  year      = {2024}
}

@article{franke2024trips,
    title={{TRIPS}: Trilinear Point Splatting for Real-Time Radiance Field Rendering},
    author={Linus Franke and Darius R{\"u}ckert and Laura Fink and Marc Stamminger},
    year = {2024},
    journal = {Computer Graphics Forum},
    volume = {43},
    number = {2},
    doi = {10.1111/cgf.15012},
}

@article{franke2025vrsplatting,
    title={{VR-Splatting}: Foveated Radiance Field Rendering via {3D} {G}aussian Splatting and Neural Points},
    author={Linus Franke and Laura Fink and Marc Stamminger},
    journal = {Proc. ACM Comput. Graph. Interact. Tech.},
    year = {2025},
    publisher = {Association for Computing Machinery},
    address = {New York, NY, USA},
    volume = {8},
    number = {1},
    month = may,
    articleno = {18},
    url = {https://doi.org/10.1145/3728302},
    doi = {10.1145/3728302}
}

@article{hahlbohm2025htgs,
  author    = {Hahlbohm, Florian and Friederichs, Fabian and Weyrich, Tim and Franke, Linus and Kappel, Moritz and Castillo, Susana and Stamminger, Marc and Eisemann, Martin and Magnor, Marcus},
  title     = {Efficient Perspective-Correct {3D} {G}aussian Splatting Using Hybrid Transparency},
  journal   = {Computer Graphics Forum},
  volume    = {44},
  number    = {2},
  year      = {2025},
  doi       = {10.1111/cgf.70014}
}

@article{radl2024stopthepop,
  title={{StopThePop}: Sorted {G}aussian Splatting for View-Consistent Real-Time Rendering},
  author={Radl, Lukas and Steiner, Michael and Parger, Mathias and Weinrauch, Alexander and Kerbl, Bernhard and Steinberger, Markus},
  journal={ACM Transactions on Graphics},
  volume={43},
  number={4},
  year={2024}
}

@inproceedings{deutsch2026ppispphysicallyplausiblecompensationcontrol,
  title     = {{PPISP}: Physically-Plausible Compensation and Control of Photometric Variations in Radiance Field Reconstruction},
  author    = {Deutsch, Isaac and Mo{\"e}nne-Loccoz, Nicolas and State, Gavriel and Gojcic, Zan},
  booktitle = {IEEE/CVF Conference on Computer Vision and Pattern Recognition (CVPR)},
  pages     = {7289--7298},
  year      = {2026}
}

@inproceedings{pan2024glomap,
    author={Pan, Linfei and Baráth, Dániel and Pollefeys, Marc and Sch\"{o}nberger, Johannes Lutz},
    title={Global Structure-from-Motion Revisited},
    booktitle={European Conference on Computer Vision (ECCV)},
    year={2024},
}

@inproceedings{meuleman2023progressively,
  title={Progressively Optimized Local Radiance Fields for Robust View Synthesis},
  author={Meuleman, Andreas and Liu, Yu-Lun and Gao, Chen and Huang, Jia-Bin and Kim, Changil and Kim, Min H and Kopf, Johannes},
  booktitle={IEEE/CVF Conference on Computer Vision and Pattern Recognition (CVPR)},
  year={2023}
}

@inproceedings{liu2026universal,
  title={Universal {B}eta Splatting},
  author={Liu, Rong and Gao, Zhongpai and Planche, Benjamin and Chen, Meida and Nguyen, Van Nguyen and Zheng, Meng and Choudhuri, Anwesa and Chen, Terrence and Wang, Yue and Feng, Andrew and others},
  booktitle={International Conference on Learning Representations (ICLR)},
  year={2026}
}

@inproceedings{yu2024mip,
  title={{Mip-Splatting}: Alias-free {3D} {G}aussian Splatting},
  author={Yu, Zehao and Chen, Anpei and Huang, Binbin and Sattler, Torsten and Geiger, Andreas},
  booktitle={IEEE/CVF Conference on Computer Vision and Pattern Recognition (CVPR)},
  year={2024}
}

@inproceedings{huang2025deformable,
  title={Deformable Radial Kernel Splatting},
  author={Huang, Yi-Hua and Lin, Ming-Xian and Sun, Yang-Tian and Yang, Ziyi and Lyu, Xiaoyang and Cao, Yan-Pei and Qi, Xiaojuan},
  booktitle={IEEE/CVF Conference on Computer Vision and Pattern Recognition (CVPR)},
  year={2025}
}

@article{ruckert2022adop,
  title={{ADOP}: Approximate Differentiable One-Pixel Point Rendering},
  author={R{\"u}ckert, Darius and Franke, Linus and Stamminger, Marc},
  journal={ACM Transactions on Graphics},
  volume={41},
  number={4},
  year={2022}
}

@inproceedings{aliev2020neural,
  title={Neural Point-Based Graphics},
  author={Aliev, Kara-Ali and Sevastopolsky, Artem and Kolos, Maria and Ulyanov, Dmitry and Lempitsky, Victor},
  booktitle={European Conference on Computer Vision (ECCV)},
  year={2020}
}

@inproceedings{harrer2023inovis,
  title={{Inovis}: Instant Novel-View Synthesis},
  author={Harrer, Mathias and Franke, Linus and Fink, Laura and Stamminger, Marc and Weyrich, Tim},
  booktitle={SIGGRAPH Asia 2023 Conference Papers},
  year={2023}
}

@article{yu2024gaussian,
  title={Gaussian opacity fields: Efficient adaptive surface reconstruction in unbounded scenes},
  author={Yu, Zehao and Sattler, Torsten and Geiger, Andreas},
  journal={ACM Transactions on Graphics (ToG)},
  volume={43},
  number={6},
  pages={1--13},
  year={2024},
  publisher={ACM, New York, NY, USA}
}

@inproceedings{sturm2012benchmark,
  title={A Benchmark for the Evaluation of {RGB-D} {SLAM} Systems},
  author={Sturm, J{\"u}rgen and Engelhard, Nikolas and Endres, Felix and Burgard, Wolfram and Cremers, Daniel},
  booktitle={IEEE/RSJ International Conference on Intelligent Robots and Systems (IROS)},
  year={2012}
}
}

\clearpage
\appendix

\newcommand{\suppltodo}[1]{\textcolor{red}{[#1]}}
\newcommand{\supplinput}[1]{\IfFileExists{#1}{\input{#1}}{%
    \fbox{\parbox{0.9\linewidth}{\centering\texttt{\detokenize{#1}} pending}}}}

\section{Derivation of the closed-form merge}
\label{sec:supp-merge-derivation}
\paragraph{(a)
  Problem formulation.
}
Given children $(o_i,\mu_i,\Sigma_i,b_i)$, we seek one parent $(o^\star,\mu^\star,\Sigma^\star,b^\star)$ that summarizes their combined density.
We use an encode--decode construction: represent the mixture by its mass, mean, covariance, and kurtosis, then invert these statistics back into the beta family.
The required ingredients are: closed-form moments for one primitive (b--c), the motivation for moment matching (d), closed-form mixture moments (e--f), and the resulting merge algorithm (g), followed by validation (h).

\paragraph{(b)
  A primitive in whitened coordinates.
}
In whitened coordinates $\mathbf{w}=\Sigma^{-1/2}(\mathbf{x}-\mu)$, the kernel is spherical: $o(1-\|\mathbf{w}\|^2/9)^\beta$.
Its radial distribution therefore contains all required information.
Since shell probability is proportional to density $\times$ the shell area, $p(r)\propto r^2(1-r^2/9)^\beta$ on $[0,3]$.
With $x=r^2/9$, as in \mainref{eq:beta_kernel}, this becomes a $\mathrm{Beta}(\tfrac32,\beta+1)$ distribution.

Hence all required radial moments are available in closed form:
\begin{equation}
  \mathbb{E}[x] = \frac{1.5}{\beta+2.5},
  \qquad
  \mathbb{E}[x^2] = \frac{15/4}{(\beta+2.5)(\beta+3.5)},
  \label{eq:tex-betamom}
\end{equation}
with no numerical integration.

\paragraph{(c)
  Moments of one primitive.
}
\emph{Mass.}
Integrating the kernel gives
\begin{equation}
  Z = 54\pi\, \mathrm{B}(\tfrac32,\beta+1)\,\sqrt{\det\Sigma}\; o .
  \label{eq:tex-mass}
\end{equation}
This is opacity $\times$ support volume $\times$ a $\beta$-dependent fill factor.

\emph{Covariance.}
Spherical symmetry gives
\begin{equation}
  \xi(\beta) = 3\,\mathbb{E}[x] = \frac{4.5}{\beta+2.5},
  \qquad
  \mathrm{Cov} = \xi(\beta)\,\Sigma .
  \label{eq:tex-shrink}
\end{equation}
Thus $\Sigma$ parameterizes the support, while $\xi(\beta)$ converts it to statistical spread.
In particular, $\xi(0)=1.8$ for the uniform radius-$3$ ball and $\xi(\beta)\to0$ as $\beta\to\infty$.

\emph{Kurtosis.}
The Mardia kurtosis $\kappa=\mathbb{E}[(\mathbf{z}^\top\mathbf{z})^2]$, after standardization by the true covariance, is invariant to position, scale, and orientation.
For this family it therefore depends only on $\beta$:
\begin{equation}
  \kappa(\beta) = 15\,\frac{\beta+2.5}{\beta+3.5}.
  \label{eq:tex-kurt}
\end{equation}
It ranges from $\kappa(0)=75/7$ for the uniform ball to $\kappa(\infty)=15$ for the Gaussian limit.
Since it is strictly
increasing, it admits the inverse
\begin{equation}
  \beta = \frac{7\kappa - 75}{2\,(15-\kappa)},
  \qquad
  b = \ln(\beta/4).
  \label{eq:tex-binv}
\end{equation}
Thus kurtosis uniquely decodes the profile shape within the family.

\begin{table*}[t]
  \centering\small\setlength{\tabcolsep}{4pt}
  \caption{\textbf{Per-stage runtime}, in milliseconds (CUDA-synchronized timers, so stage costs are exact but totals are overestimated).
    Image load and feature extraction run once per streamed frame; all other stages run per registered keyframe (loop closure rarely $1-3\times$).
  }
  \label{tab:suppl-runtime}
  \begin{tabular}{lrrrrr}
    \toprule
    Stage (ms/call)      & \textsc{Garden} & \textsc{fr1/desk} & \textsc{Truck} & \textsc{University2} & \textsc{SmallCity} \\
    \midrule
    Image load, features & 13.4            & 11.8              & 13.4           & 11.4                 & 13.2               \\
    Pose estimation      & 119.2           & 149.2             & 117.6          & 114.6                & 119.7              \\
    Loop closure         & 101.3           & --                & 164.1          & --                   & --                 \\
    Primitive spawning   & 47.1            & 49.3              & 46.0           & 44.6                 & 36.4               \\
    Optimization burst   & 170.6           & 146.5             & 129.2          & 174.9                & 329.6              \\
    Hierarchy update     & 2.7             & 2.7               & 3.0            & 41.4                 & 56.5               \\
    \midrule
    Registered keyframes & 180             & 159               & 228            & 347                  & 369                \\
    Primitives (final)   & 2.24M           & 0.33M             & 1.23M          & 2.44M                & 1.32M              \\
    Total time (m:ss)    & 1:02            & 1:05              & 1:09           & 2:07                 & 3:47               \\
    \bottomrule
  \end{tabular}
\end{table*}

\paragraph{(d)
  Why moment matching.
}
For Gaussians, the M-projection $\arg\min_q\mathrm{KL}(p\,\|\,q)$ reduces to moment matching.
For compact beta kernels, however, KL is unsuitable because it is infinite wherever $p>0$ but $q=0$.
A child extending even infinitesimally beyond the candidate parent's support therefore dominates the divergence despite carrying negligible mass.
Moment matching avoids this support pathology while retaining the Gaussian-limit interpretation.

\paragraph{(e)
  Mixture mass, mean, and covariance.
}
Let $w_i\propto Z_i$ with $\sum_iw_i=1$.
The mixture mean is \[ \mu^\star=\sum_i w_i\mu_i.
\]
By the law of total covariance,
\begin{equation}
  \Sigma_{\mathrm{mix}} = \sum_i w_i \big(\xi(\beta_i)\Sigma_i+\delta_i\delta_i^\top\big), \qquad \delta_i=\mu_i-\mu^\star .
  \label{eq:tex-totalcov}
\end{equation}
Crucially, each child contributes its true covariance $\xi(\beta_i)\Sigma_i$, not its support matrix $\Sigma_i$.

\paragraph{(f)
  Mixture kurtosis.
}
The mixture kurtosis is evaluated with respect to the mixture covariance $\Sigma_{\mathrm{mix}}$:
\begin{equation*}
  \kappa_{\mathrm{mix}}=\sum_iw_i \mathbb{E}_i[(\mathbf{z}^\top\mathbf{z})^2],
\end{equation*}
where $\mathbf{z}^\top\mathbf{z}= (\mathbf{x}-\mu^\star)^\top \Sigma_{\mathrm{mix}}^{-1}(\mathbf{x}-\mu^\star)$.

Write a sample from child $i$ as $\mathbf{x}=\mu_i+\Sigma_i^{1/2}\mathbf{s}$, where $\mathbf{s}$ is isotropic with radial moments $m_2=9\mathbb{E}_i[x]$ and $m_4=81\mathbb{E}_i[x^2]$.
Define
\begin{align}
  a_i = \delta_i^\top \Sigma_{\mathrm{mix}}^{-1}\delta_i, \\
  v_i = \Sigma_i^{1/2}\Sigma_{\mathrm{mix}}^{-1}\delta_i, \\
  M_i = \Sigma_i^{1/2}\Sigma_{\mathrm{mix}}^{-1}\Sigma_i^{1/2},
  \label{eq:tex-avm}
\end{align}
so that
\begin{equation}
  \mathbf{z}^\top\mathbf{z}
  = a_i + 2\,v_i^\top\mathbf{s} + \mathbf{s}^\top M_i\,\mathbf{s}.
\end{equation}

Odd terms vanish by symmetry.
Isotropy gives
$\mathbb{E}[\mathbf{s}\mathbf{s}^\top]=\tfrac{m_2}{3}I$ and
\begin{equation}
  \mathbb{E}\big[(\mathbf{s}^\top M\mathbf{s})^2\big]
  = \tfrac{m_4}{15}
  \big((\operatorname{tr}
  M)^2+2\operatorname{tr}M^2\big).
  \label{eq:tex-iso4}
\end{equation}
Therefore
\begin{equation}
  \begin{split}
    \mathbb{E}_i\big[(\mathbf{z}^\top\mathbf{z})^2\big]
    = a_i^2
    + \tfrac{4m_2}{3}\,\|v_i\|^2          \\
    + \tfrac{m_4}{15}\big((\operatorname{tr}
    M_i)^2 + 2\operatorname{tr}M_i^2\big) \\ + \tfrac{2a_i m_2}{3} \operatorname{tr}M_i .
  \end{split}
  \label{eq:tex-perchild}
\end{equation}

Thus each contribution requires only traces of products of $3{\times}3$ matrices and is readily batched on the GPU.

\paragraph{Out-of-band mixtures.}
A mixture of beta kernels can fall outside the family's kurtosis band.
This happens on both sides: well-separated children give lighter tails than any beta kernel, while heavy-tailed groups exceed $15$.
Since $\kappa$ is scale-invariant, shrinking a group to sub-pixel size does not change this.
We restrict the parent to $b^\star\in[-4,4]$ and clip $\kappa_{\mathrm{mix}}$ to $\hat\kappa$ in the corresponding band $[\kappa(4e^{-4}),\kappa(4e^{4})]$.
Because $\kappa$ is monotone, this selects the closest representable profile and keeps $\beta^\star$ finite.
Parent opacity is capped at the rasterizer's limit of $0.99$.
Clipping is frequent on large scenes, yet the merge still clearly outperforms the opacity-weighted baseline (\mainref{tab:abl-flagship}).

\paragraph{(g)
  End-to-end merge.
}
The merge proceeds as follows:
\begin{enumerate}
  \item Compute $Z_i$ from \eqref{eq:tex-mass} and normalize to $w_i$.
  \item Compute $\mu^\star=\sum_i w_i\mu_i$.
  \item Compute $\Sigma_{\mathrm{mix}}$ from \eqref{eq:tex-totalcov}.
  \item Evaluate \eqref{eq:tex-perchild} and accumulate
        $\kappa_{\mathrm{mix}}$.
  \item Clip $\kappa_{\mathrm{mix}}$ to the band of $b^\star\in[-4,4]$.
  \item Decode $\beta^\star$ and $b^\star$ using \eqref{eq:tex-binv}.
  \item Recover $\Sigma^\star=\Sigma_{\mathrm{mix}}/\xi(\beta^\star)$ and conserve mass through
        $o^\star=\sum_iZ_i/Z(\beta^\star,\Sigma^\star)$, capped at $0.99$.
\end{enumerate}

The procedure requires no sampling or iterative fitting.
Its only nontrivial linear algebra consists of symmetric $3{\times}3$ square roots, inverses, and the final decomposition of $\Sigma^\star$, all handled by batched $3{\times}3$ eigendecompositions.

\paragraph{(h)
  Stress tests.
}
\emph{Coincident children}.
For $N$ identical children, $\delta_i=0$ and the mixture moments reduce to those of the child itself.
Consequently the decoder recovers $\beta^\star=\beta$, $\Sigma^\star=\Sigma$, and $o^\star=N \cdot o$, which can exceed $1$ and is then capped.

\emph{Out-of-band groups.}
The kurtosis range also acts as a validity certificate.
For four near-point children at the vertices of a regular tetrahedron, $\Sigma_i\to0$ leaves only the center-offset terms.
Symmetry gives $\Sigma_{\mathrm{mix}}=\tfrac{R^2}{3}I$ and $a_i=3$, yielding $\kappa_{\mathrm{mix}}=9<75/7$.
Since this lies outside the beta family's kurtosis range, the merge is clipped to the flattest admissible profile.
At the opposite boundary, mixtures with sufficiently heavy tails satisfy $\kappa_{\mathrm{mix}}\geq15$ and clamp to $b^\star=4$.

\emph{Numerical validation.}
Across 200 randomized four-child merges, Monte-Carlo estimates using $1.5$M samples per merge agree with the closed forms to $0.19\%$ in kurtosis and $0.13\%$ in covariance.

\section{Implementation Details \& Startup}
\label{sec:suppl-impl}

The system is implemented in PyTorch~2.11/CUDA~12.8, with performance-critical stages implemented as custom GPU kernels.
Feature extraction uses XFeat~\cite{potje2024xfeat}, matching combines mutual nearest neighbors with LighterGlue~\cite{lindenberger2023lightglue}, retrieval uses MixVPR~\cite{alibey2023mixvpr}, and monocular depth priors use Depth Anything V2~\cite{yang2024depthanythingv2}.
The pose-estimation and mapping stages follow the backbone of i3DGS~\cite{meuleman2026immediate}, including P3P/RANSAC, triangulation, and local bundle adjustment.

Rasterization uses our beta-kernel implementation of a FasterGS-style fused rasterizer~\cite{hahlbohm2026fastergs}.
Forward quantities are recomputed during the backward pass, and the sparse Adam update and analytic regularizer gradients are applied directly to the optimizer state without materializing primitive gradients.
Spherical-beta colour parameters are stored compactly for the fused implementation, while exported models retain the standard SH-width layout for compatibility.

Each keyframe performs one $30$-iteration optimization burst.
The bootstrap uses $8$ keyframes, local BA operates on the $5$ most similar keyframes, and the LoD hierarchy is rebuilt every $10$ keyframes.
Inactive fine levels and keyframes beyond the active window are offloaded to host memory.
All reported timings include the online pose estimation, spawning, optimization, and hierarchy stages.

\section{Runtime and Latency}
\label{sec:suppl-runtime}

\begin{table}[t]
  \centering\small\setlength{\tabcolsep}{3.5pt}
  \caption{\textbf{Time to first image.}
    Time from the first streamed frame to the first renderable reconstruction, including bootstrap, mini-BA, and the first optimization burst.
    ``Frames used'' denotes the number of streamed frames consumed by
    bootstrap.
  }
  \label{tab:suppl-firstimage}
  \begin{tabular}{lcccc}
    \toprule
                         & 1st image & Frames & Boot. & Mean latency/ \\
    Scene                & (s)       & used   & KFs   & frame (ms)    \\
    \midrule
    \textsc{Garden}      & 1.2       & 8      & 8     & 336           \\
    \textsc{fr1/desk}    & 1.5       & 17     & 8     & 106           \\
    \textsc{Truck}       & 1.0       & 10     & 8     & 276           \\
    \textsc{University2} & 1.2       & 9      & 8     & 176           \\
    \textsc{SmallCity}   & 1.4       & 9      & 8     & 237           \\
    \bottomrule
  \end{tabular}
\end{table}

\paragraph{Per-keyframe runtime.}
\cref{tab:suppl-runtime} reports the runtime of the main online stages.
Optimization is the dominant cost ($129$--$330$\,ms per keyframe), followed by pose estimation and primitive spawning.
Hierarchy updates remain small on the bounded scenes and become more significant for \textsc{University2} and \textsc{SmallCity}.
Frames that are not registered as keyframes incur only image processing and matching, resulting in lower average latency on TUM and StaticHikes.
Using the fused rasterizer reduces the optimization cost on \textsc{Garden} from $287$\,ms to $170$\,ms per keyframe and total runtime from $84$\,s to $62$\,s.

\paragraph{Time to first image.}
\cref{tab:suppl-firstimage} measures the time from the first streamed frame
to the first renderable reconstruction.
Across the evaluated scenes, the first image is available within $1.0$--$1.5$\,s, after an $8$-keyframe bootstrap, mini-BA, and first $30$-iteration optimization burst.

\begin{table}[t]
  \centering\small\setlength{\tabcolsep}{4pt}
  \caption{\textbf{Fused rasterization step profile} on a converged
    \textsc{Garden} reconstruction (2.2M primitives, $1297{\times}840$).
    The fused backward integrates the sparse Adam update in-kernel, so the per-primitive stages carry the optimizer's memory traffic: running the preprocess backward with plain gradient stores instead of the fused Adam update halves it ($0.66$ vs.
    \ $1.31$\,ms) --- i.e.\ half that kernel is the optimizer.
    End-to-end, the fused path reduces the 30-iteration burst from $287$ to $170$\,ms per keyframe (\cref{tab:suppl-runtime}).
  }
  \label{tab:suppl-raster}
  \begin{tabular}{lrr}
    \toprule
    Stage                                     & ms   & share \\
    \midrule
    blend backward                            & 1.67 & 37\%  \\
    preprocess backward (fused Adam)          & 1.31 & 29\%  \\
    blend forward                             & 0.85 & 19\%  \\
    preprocess forward                        & 0.36 & 8\%   \\
    sort / instancing                         & 0.37 & 8\%   \\
    \midrule
    \quad preprocess backward (no fused Adam) & 0.66 &       \\
    \bottomrule
  \end{tabular}
\end{table}

\paragraph{Fused optimization.}
\cref{tab:suppl-raster} profiles one fused optimization iteration.
Integrating the sparse Adam update removes the separate gradient write and optimizer read, substantially reducing the optimization cost.
On \textsc{Garden}, this reduces the optimization burst from $287$ to $170$\,ms per keyframe (\cref{tab:suppl-runtime}).

\section{Additional Results}
\label{sec:suppl-additional-res}
\begin{table*}[t]
  \begin{minipage}[t]{0.69\textwidth}
    \centering\footnotesize\setlength{\tabcolsep}{0.25pt}
    \captionof{table}{\textbf{Low-resolution protocol} (TUM resized to $448{\times}336$, MipNeRF360 and StaticHikes to 640\,px width). $^\dagger$ from i3DGS~\cite{meuleman2026immediate}, with times roughly scaled to our setup. \textbf{Bold}: best.
    }\vspace{-2.5mm}
    \label{tab:res-half}
    \footnotesize
    \begin{tabular}{l|cccc|cccc|cccc}
      \toprule
                                                              &
      \multicolumn{4}{c|}{\textsc{TUM}}                       &
      \multicolumn{4}{c|}{\textsc{MipNeRF360} \textit{(otf)}} &
      \multicolumn{4}{c}{\textsc{StaticHikes}}                                                                                                                                                                                                                                        \\
      {}                                                      & PSNR$^\uparrow$ & SSIM$^\uparrow$    & LPIPS$^\downarrow$ & Time$^\downarrow$ &
      PSNR$^\uparrow$                                         & SSIM$^\uparrow$ & LPIPS$^\downarrow$ & Time$^\downarrow$  &
      PSNR$^\uparrow$                                         & SSIM$^\uparrow$ & LPIPS$^\downarrow$ & Time$^\downarrow$                                                                                                                                                              \\
      \midrule
      DROID-Splat$^\dagger$                                   & 19.49           & 0.721              & 0.325              & $\sim$5min        & 25.87          & 0.776          & 0.258          & $\sim$9min     & 19.65          & 0.470          & 0.506          & $\sim$8min     \\
      CF-3DGS$^\dagger$                                       & 15.05           & 0.578              & 0.405              & $\sim$1h          & 13.52          & 0.295          & 0.621          & $\sim$1h       & 15.21          & 0.301          & 0.560          & $\sim$7h       \\
      OtF NVS$^\dagger$                                       & 22.45           & 0.815              & 0.225              & $\sim$50s         & 25.80          & 0.834          & 0.182          & $\sim$40s      & 21.93          & 0.673          & \textbf{0.270} & $\sim$
      1min                                                                                                                                                                                                                                                                            \\
      ARTDECO                                                 & 23.03           & 0.662              & 0.343              & 04:56             & 22.97          & 0.841          & 0.210          & 14:07          & 21.28          & 0.555          & 0.380          & 13:05          \\
      i3DGS                                                   & 22.51           & 0.836              & 0.189              & \textbf{00:58}    & 27.16          & 0.849          & 0.165          & \textbf{00:43} & \textbf{22.24} & \textbf{0.683} & 0.297          & \textbf{01:04} \\
      \midrule
      Ours                                                    & \textbf{25.14}  & \textbf{0.872}     & \textbf{0.148}     & 01:26             & \textbf{27.73} & \textbf{0.867} & \textbf{0.144} & 00:49          & 22.17          & 0.679          & 0.309          & 01:12          \\
      \bottomrule
    \end{tabular}
  \end{minipage}
  \hfill
  \begin{minipage}[t]{0.29\textwidth}
    \centering\footnotesize\setlength{\tabcolsep}{0.25pt}
    \caption{\textbf{Tiny resolution ($448{\times}448$) input.} $^\dagger$ from i3DGS~\cite{meuleman2026immediate}.
      $R$ indicates refinement following~\cite{jiang2025anysplat}.
      \textbf{Bold}: best.
    }\vspace{-2.8mm}
    \label{tab:res-448}
    \begin{tabular}{l|cccr}
      \toprule
                         & \multicolumn{4}{c}{\textsc{MipNeRF360} \textit{(otf)}}
      \\
                         & PSNR$\uparrow$                                         & SSIM$\uparrow$ & LPIPS$\downarrow$ & Time$\downarrow$ \\
      \midrule
      AnySplat$^\dagger$ & 18.05                                                  & 0.430          & 0.390             & $\sim$20s        \\
      AnySplat$^{R}$$^\dagger$ & 21.53                                                  & 0.600          & 0.300             & $\sim$2min \\
      i3DGS              & 26.51                                                  & \textbf{0.845} & 0.173             & 0:38             \\
      DA3-Gauss          & 19.88                                                  & 0.487          & 0.338             & \textbf{0:17}    \\
      \midrule
      Ours               & \textbf{26.56}                                         & 0.842          & \textbf{0.166}    & 0:46             \\
      \bottomrule
    \end{tabular}
  \end{minipage}
  \vspace{-1mm}
\end{table*}

\paragraph{Reduced-Resolution Protocols.}
\cref{tab:res-half} evaluates MipNeRF360 (\textit{otf}), TUM, and StaticHikes at reduced resolution.
We remain strongest on the image metrics for TUM and MipNeRF360, although the reduced-detail setting negates the speed advantage and degrades performance on StaticHikes scenes, where there is less sharp detail to explain.
We also surpass ARTDECO, which becomes substantially more expensive at higher resolution (StaticHikes and MipNeRF360).

\paragraph{Vision Transformer Protocol.}
In \cref{tab:res-448}, we evaluate AnySplat~\cite{jiang2025anysplat} and DepthAnything3~\cite{lin2025depthanything3} (Giant-Large-1.1, Gaussian head) under the $448{\times}448$ protocol.
Despite their feed-forward reconstruction, our method has increased reconstruction quality by at least $5$\,dB, including over the post-trained AnySplat variant.

\section{Colour Basis on DeepBlending}
\label{sec:suppl-shdb}

\begin{table}[t]
  \centering\small\setlength{\tabcolsep}{4pt}
  \caption{\textbf{Colour basis on DeepBlending.}
    Comparison of the default five zero-initialized spherical-beta lobes with degree-3 SH using the identical reconstruction pipeline.
  }
  \label{tab:suppl-shdb}
  \begin{tabular}{llccc}
    \toprule
    Scene     & Colour   & PSNR$^\uparrow$ & SSIM$^\uparrow$ & LPIPS$^\downarrow$ \\
    \midrule
    drjohnson & SB lobes & 25.67           & 0.828           & 0.315              \\
              & SH       & 25.88           & 0.832           & 0.309              \\
    \midrule
    playroom  & SB lobes & 27.28           & 0.865           & 0.276              \\
              & SH       & 27.58           & 0.868           & 0.268              \\
    \bottomrule
  \end{tabular}
\end{table}

\begin{table}[t]
  \centering\small\setlength{\tabcolsep}{4.5pt}
  \caption{\textbf{Spherical-beta lobes vs.\ SH colour, per scene.}
    $\Delta$ denotes lobes minus SH for PSNR and LPIPS.
    Both variants use the same reconstruction configuration except color representation.
  }
  \label{tab:sh-vs-lobes}
  \begin{tabular}{llccc}
    \toprule
    Dataset     & Scene       & $\Delta$PSNR & $\Delta$SSIM & $\Delta$LPIPS \\
    \midrule
    db          & drjohnson   & -0.21        & -0.004       & +0.007        \\
    db          & playroom    & -0.29        & -0.003       & +0.008        \\
    \midrule
    MipNeRF360  & garden      & +0.18        & +0.005       & -0.000        \\
    MipNeRF360  & bonsai      & +0.23        & +0.003       & -0.002        \\
    MipNeRF360  & counter     & +0.41        & +0.007       & -0.006        \\
    MipNeRF360  & bicycle     & +0.16        & +0.013       & -0.001        \\
    MipNeRF360  & kitchen     & +0.25        & +0.007       & -0.004        \\
    MipNeRF360  & room        & -0.03        & -0.004       & +0.002        \\
    MipNeRF360  & stump       & -0.16        & -0.009       & +0.014        \\
    MipNeRF360  & flowers     & +0.03        & +0.005       & +0.005        \\
    MipNeRF360  & treehill    & -0.08        & -0.001       & +0.002        \\
    \midrule
    TUM         & TUM desk    & +0.19        & +0.004       & -0.003        \\
    TUM         & TUM xyz     & +0.27        & +0.002       & +0.001        \\
    TUM         & TUM office  & +0.62        & +0.014       & -0.016        \\
    \midrule
    T\&T        & train       & +0.04        & +0.002       & +0.001        \\
    T\&T        & truck       & +0.16        & +0.005       & -0.001        \\
    \midrule
    StaticHikes & forest1     & +0.02        & +0.003       & +0.001        \\
    StaticHikes & forest2     & -0.03        & +0.001       & +0.003        \\
    StaticHikes & university2 & -0.08        & -0.002       & +0.003        \\
    \bottomrule
  \end{tabular}
\end{table}

On DeepBlending, degree-3 SH outperforms the spherical-beta (SB) lobes used in our default configuration.
\cref{tab:suppl-shdb} shows that SH improves PSNR by
$0.2-0.3$\,dB.
The per-scene comparison shows that this behavior is largely specific to this dataset; with \textsc{office}, \textsc{counter}, and \textsc{TUM/xyz} showing strong improvements with SB (\cref{tab:sh-vs-lobes}).

\section{Kernel Shape vs.
  \ Footprint}
\label{sec:suppl-kernel}

\begin{figure}[t]
  \centering
  \includegraphics[width=\linewidth]{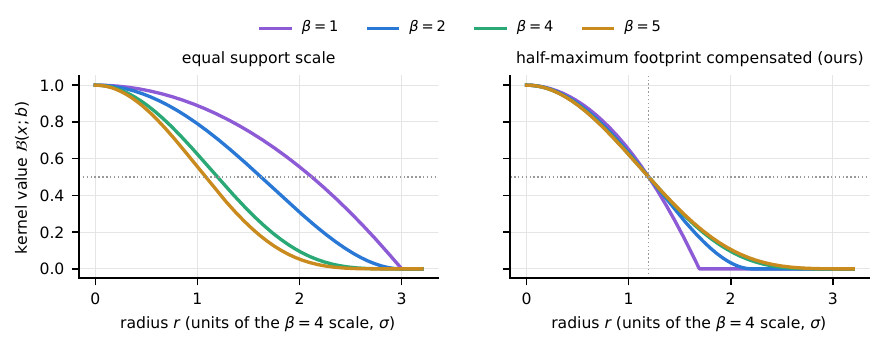}
  \caption{\textbf{Half-maximum scale compensation.}
    Without compensation, lowering $\beta$ broadens the kernel; with the proposed compensation, all profiles share the same half-maximum radius, isolating shape changes from footprint changes.
  }
  \label{fig:suppl-footprint}
\end{figure}

\cref{fig:suppl-footprint} verifies the scale compensation used whenever
$\beta$ changes.
Without it, a $\beta=1$ kernel is about $1.8\times$ wider at half maximum than the $\beta=4$ reference, also increasing its footprint area by roughly $3\times$.
The proposed compensation preserves the half-maximum radius, allowing spawning, solidification, and merging to change solidity without changing apparent size.

\section{Pose Accuracy on TUM}
\label{sec:suppl-ate}

\begin{table}[t]
  \centering\small\setlength{\tabcolsep}{1pt}
  \caption{\textbf{TUM pose accuracy.}
    Our method and i3DGS evaluated against the same COLMAP pseudo-ground truth using aligned trajectories.
    Translation errors are reported as a percentage of trajectory extent and rotation errors in degrees; results are averaged over multiple runs.
  }
  \label{tab:suppl-ate}
  \begin{tabular}{lccccc}
    \toprule
    Scene      & Method & ATE(\%)$^\downarrow$ & R.APE($^\circ$)$^\downarrow$ & T.RPE(\%)$^\downarrow$ & R.RPE($^\circ$)$^\downarrow$ \\
    \midrule
    fr1/desk   & Ours   & 0.55                 & 1.90                         & 0.344                  & 0.644                        \\
               & i3DGS  & \textbf{0.47}        & \textbf{1.42}                & \textbf{0.314}         & \textbf{0.548}               \\
    \midrule
    fr2/xyz    & Ours   & \textbf{0.33}        & \textbf{0.41}                & \textbf{0.178}         & \textbf{0.128}               \\
               & i3DGS  & 0.37                 & 0.42                         & 0.185                  & 0.133                        \\
    \midrule
    fr3/office & Ours   & \textbf{0.32}        & \textbf{0.94}                & \textbf{0.127}         & \textbf{0.326}               \\
               & i3DGS  & 0.38                 & 1.02                         & 0.150                  & 0.347                        \\
    \bottomrule
  \end{tabular}
\end{table}

\begin{table}[t]
  \centering\small\setlength{\tabcolsep}{1pt}
  \caption{\textbf{Conditioning the \textsc{fr1/desk} bootstrap.}
    Three bootstrap variants under the same scoring protocol.
    The spread selector improves pose accuracy and PSNR at the cost of $1.1$\,s additional latency; the parallax floor degrades accuracy.
  }
  \label{tab:suppl-posefix}
  \begin{tabular}{lccccc}
    \toprule
    Bootstrap                   & ATE (\%)$^\downarrow$ & R.
    APE ($^\circ$)$^\downarrow$ & PSNR$^\uparrow$       & f.i.
    (s)                         & $f$                                              \\
    \midrule
    ordered first-8             & 0.55                  & 1.90 & 23.93 & 1.5 & 519 \\
    ordered + parallax          & 1.78                  & 3.17 & 23.67 & 1.5 & 527 \\
    spread, pool 30             & 0.51                  & 1.21 & 24.20 & 2.6 & 511 \\
    \bottomrule
  \end{tabular}
\end{table}

\cref{tab:suppl-ate} compares our pose estimation with i3DGS on the three
TUM sequences using the same trajectories and COLMAP pseudo-ground truth.
On \textsc{fr2/xyz} and \textsc{fr3/office}, our trajectories are slightly more accurate than those of i3DGS on all four metrics.
On \textsc{fr1/desk}, i3DGS remains marginally ahead (ATE $0.55$ vs.
\ $0.47\%$,
rotation APE $1.90^\circ$ vs.\ $1.42^\circ$).
This sequence is the hardest of the three for our bootstrap: it is strongly motion-blurred, and the first eight frames, which we use to reach the first image quickly, provide little parallax.

\cref{tab:suppl-posefix} shows how sensitive \textsc{fr1/desk} is to the
choice of bootstrap frames.
All three variants converge to a focal length close to the calibrated value ($511$--$527$ vs.
\ ${\approx}517$), so the differences come from the trajectory
rather than the intrinsics.
Enforcing a parallax floor on the ordered bootstrap is counterproductive (ATE $1.78\%$).
Selecting a spread-maximizing subset from the first 30 frames reduces ATE to $0.51\%$ and rotation APE to $1.21^\circ$, on par with or better than i3DGS, and also improves PSNR by $0.27$\,dB.
Its only cost is $1.1$\,s of additional time to first image.
Our default ordered bootstrap therefore trades a small amount of pose accuracy on \textsc{fr1/desk} for lower latency; the underlying pose pipeline is on par with i3DGS, and a wider bootstrap pool closes the remaining gap when the extra second of latency is acceptable.

\section{Confidence}

\subsection{Solidity Controller Sweep and Sensitivity}
\label{sec:suppl-solidity}

\begin{figure}[]
  \centering
  \includegraphics[width=\linewidth]{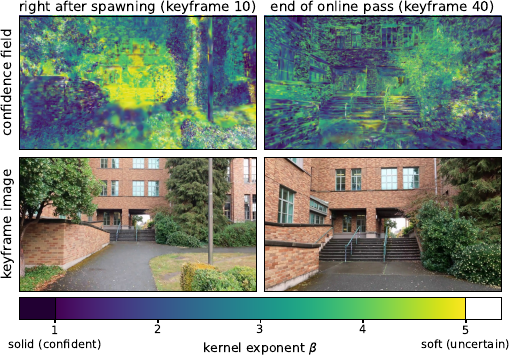}
  \vspace{-6mm}
  \caption{\textbf{Online confidence field on a walking capture}
    (\textsc{University2}), analogous to \mainref{fig:conf_field}:
    renders over their per-primitive shape field ($\beta$, warm =
    soft).
    As the walk advances, accumulated multi-view evidence solidifies the traversed corridor of the scene behind the camera, while freshly entered and peripheral regions remain soft probes --- the field marks the reconstruction frontier of an unbounded capture live, at no quality cost.
  }
  \label{fig:suppl-conf-u2}
\end{figure}

\cref{tab:suppl-solidity} evaluates the controller step $\Delta b$ on
\textsc{Garden}.
Performance is stable for steps from $0.1$ to $0.8$.
Although the first two slightly degrade quality, the optimization is stable with regards to this hyperparameter.
The number of solidification events remains similar across settings, indicating that the evidence gates primarily determine which primitives are solidified, while $\Delta b$ controls the update rate.
We therefore use $\Delta b=0.4$ as a stable operating point.
The resulting shape field provides the confidence visualization shown in \mainref{fig:conf_field}; \cref{fig:suppl-conf-u2} shows the same behavior on \textsc{University2}.

\subsection{Spawn Confidence}
\label{sec:suppl-spawnconf}

\begin{figure*}[h]
  \centering
  \IfFileExists{figures/suppl/spawnconf_u2.pdf}{%
    \includegraphics[width=\textwidth]{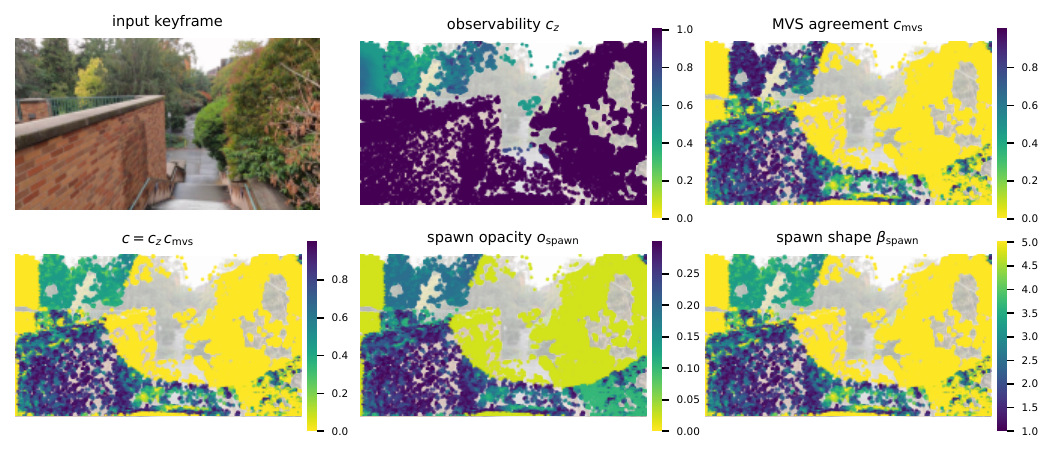}}{%
    \fbox{spawnconf figure pending}}\vspace{-3mm}
  \caption{\textbf{Spawn-confidence signals on \textsc{University2}.}
    Primitive locations are colored by geometric observability $c_z$, MVS--monocular agreement $c_{\mathrm{mvs}}$, and their product $c$.
    The two signals fail in different regions: distant trees have low observability, while dense foliage has poor depth agreement despite sufficient baseline.
    Their product therefore assigns solid, opaque initialization to well-constrained surfaces and soft initialization to ambiguous regions.
  }
  \label{fig:suppl-spawnconf}
\end{figure*}

\cref{fig:suppl-spawnconf} visualizes the two spawn-confidence signals on
\textsc{University2}.
Geometric observability suppresses poorly constrained far-field depth, while MVS--monocular agreement rejects ambiguous, occlusion-dominated regions.
Their product therefore assigns high confidence to nearby, well-verified surfaces and low confidence to uncertain geometry.

The correlation is also quantitative.
Among $12.7$k MVS-supported \textsc{Garden} spawn candidates with held-out COLMAP depth $z_{\mathrm{GT}}$, we label a candidate depth-accurate when $|\log(z/z_{\mathrm{GT}})|<0.05$ ($5\%$ relative error).
The MVS confidence ranks an accurate candidate above an inaccurate one with AUC $0.92$, whereas $c_z$ is largely uninformative on \textsc{Garden} because the baseline saturates its observability score.
On \textsc{University2}, where far-field content is present, $c_z$ provides the complementary geometric signal.
\begin{table}[t]
  \centering\small\setlength{\tabcolsep}{2pt}
  \caption{\textbf{Relocation design choices} on \textsc{fr2/xyz}.
    Each row removes one component of the error-driven relocation.
  }
  \label{tab:suppl-reloc}
  \begin{tabular}{lccc}
    \toprule
    Variant                                  & PSNR$^\uparrow$ & SSIM$^\uparrow$ & LPIPS$^\downarrow$ \\
    \midrule
    Full (error-weighted + consistency gate) & 28.45           & 0.893           & 0.133              \\
    w/o error weighting (opacity-sampled)    & 27.89           & 0.885           & 0.148              \\
    w/o relocation                           & 28.21           & 0.890           & 0.136              \\
    \bottomrule
  \end{tabular}
\end{table}

\providecommand{\rotrow}[1]{%
  \multirow{4}{25pt}{\rotatebox[origin=c]{90}{{#1}}}}
\begin{table*}[!h]
  \setlength{\tabcolsep}{3pt}
  \caption{\label{tab:gs_benchmark_full}
    \textbf{Results on the full 3DGS benchmark.}
    Nine MipNeRF360, two Tanks\&Temples, and two Deep Blending scenes.
    We evaluate BetaSplatting and FasterGS with GLOMAP poses using sequential and exhaustive matching, and compare against i3DGS and our method.
    GLOMAP sequential requires 0:01:52 on MipNeRF360, 0:01:42 on Tanks\&Temples and fails on Deep Blending.
    GLOMAP exhaustive takes respectively 0:04:53, 0:04:22 and 0:03:42.
    The {\colorbox{firstcolor}{best}} and {\colorbox{secondcolor}{second best}} are color coded for non-exhaustive matching methods.
  }
  \vspace{-7pt}
  \setlength{\tabcolsep}{3pt}
  \footnotesize

  \centering
  \begin{tabular}{@{}c@{\hspace{-12pt}}l|cccc|cccc|cccc}\toprule
                                                    &                     &
    \multicolumn{4}{c|}{\textsc{MipNeRF360}}        &
    \multicolumn{4}{c|}{\textsc{Tanks and Temples}} &
    \multicolumn{4}{c}{\textsc{Deep Blending}}                                                                                                                                                                                                                                                          \\
    {}                                    &                     & PSNR$^\uparrow$    & SSIM$^\uparrow$   & LPIPS$^\downarrow$ & Time$^\downarrow$ &
    PSNR$^\uparrow$                                 & SSIM$^\uparrow$     & LPIPS$^\downarrow$ & Time$^\downarrow$ &
    PSNR$^\uparrow$                                 & SSIM$^\uparrow$     & LPIPS$^\downarrow$ & Time$^\downarrow$                                                                                                                                                                                      \\\hline

    \rotrow{G (\textit{exh.)}}                      & BetaSplatting (30K) & 28.09              & 0.822             & 0.222              & 0:12:34           & 22.18          & 0.818          & 0.209          & 0:11:14          & 30.49          & 0.915          & 0.250          & 0:10:11          \\

                                                    & FasterGS (30K)      & 27.66              & 0.818             & 0.251              & 0:09:01           & 20.45          & 0.781          & 0.274          & 0:07:03          & 29.78          & 0.908          & 0.306          & 0:07:14          \\
                                                    & BetaSplatting (7K)  & 25.92              & 0.750             & 0.308              & 0:06:21           & 19.72          & 0.710          & 0.347          & 0:05:36          & 26.85          & 0.861          & 0.350          & 0:04:56          \\
                                                    & FasterGS (7K)       & 25.36              & 0.740             & 0.347              & 0:05:34           & 19.07          & 0.712          & 0.373          & 0:04:48          & 28.91          & 0.891          & 0.358          & 0:04:17          \\

    \hline\hline
    \rotrow{G (\textit{seq.)}}                      & BetaSplatting (30K) & \first{}25.34      & 0.679             & 0.341              & 0:09:07           & \first{}23.31  & \first{}0.831  & \first{}0.190  & 0:08:27          & $\times$       & $\times$       & $\times$       & $\times$         \\
                                                    & FasterGS (30K)      & 24.96              & 0.679             & 0.362              & 0:05:29           & 22.17          & 0.802          & 0.250          & 0:04:21          & $\times$       & $\times$       & $\times$       & $\times$         \\
                                                    & BetaSplatting (7K)  & 23.62              & 0.636             & 0.402              & 0:03:17           & 20.81          & 0.733          & 0.321          & 0:02:55          & $\times$       & $\times$       & $\times$       & $\times$         \\
                                                    & FasterGS (7K)       & 23.96              & 0.654             & 0.413              & 0:02:29           & 20.79          & 0.742          & 0.343          & 0:02:10          & $\times$       & $\times$       & $\times$       & $\times$         \\

    \hline
    \multirow{2}{20pt}                              & i3DGS               & 24.75              & \second{}0.721    & \second{}0.280     & \second{}0:01:59  & 22.08          & 0.783          & 0.226          & \second{}0:01:25 & \first{}27.16  & \first{}0.856  & \first{}0.285  & \second{}0:02:03 \\
    \cline{2-14}
                                                    & Ours                & \second{}25.19     & \first{}0.731     & \first{}0.264      & \first{}0:01:17   & \second{}23.00 & \second{}0.803 & \second{}0.200 & \first{}0:01:18  & \second{}26.49 & \second{}0.847 & \second{}0.295 & \first{}0:01:22  \\
    \bottomrule
  \end{tabular}
\end{table*}

\section{Relocation Details}
\label{sec:suppl-reloc}
\ifdefined\arxivversion\begin{table}[!tb]
  \centering\small\setlength{\tabcolsep}{4pt}
  \caption{\textbf{Unordered input.}
    Identical configuration.
  }
  \label{tab:suppl-shuffle}
  \begin{tabular}{llcccc}
    \toprule
    Scene             & Order    & PSNR$^\uparrow$ & SSIM$^\uparrow$ & LPIPS$^\downarrow$ \\
    \midrule
    \textsc{Garden}   & ordered  & 25.82           & 0.786           & 0.187              \\
                      & shuffled & 25.34           & 0.774           & 0.191              \\
    \midrule
    \textsc{Counter}  & ordered  & 26.98           & 0.860           & 0.222              \\
                      & shuffled & 27.54           & 0.877           & 0.204              \\
    \midrule
    \textsc{Bonsai}   & ordered  & 29.61           & 0.909           & 0.204              \\
                      & shuffled & 27.11           & 0.850           & 0.261              \\
    \midrule
    \textsc{fr1/desk} & ordered  & 23.93           & 0.844           & 0.209              \\
                      & shuffled & 23.09           & 0.870           & 0.179              \\
    \bottomrule
  \end{tabular}
\end{table}
\fi
\begin{table}[!tb]
  \centering\footnotesize\setlength{\tabcolsep}{1pt}
  \caption{\textbf{Method constants and their provenance.}
    Values are grouped by spawning, solidity control, relocation, optimization/rasterization, and the hierarchy.
    \emph{Derived} quantities follow from other parameters,
    \emph{structural} values are fixed by the kernel construction, and
    \emph{inherited} values are taken unchanged from prior work.}
  \label{tab:suppl-hparams}
  \resizebox{\linewidth}{!}{%
  \begin{tabular}{llll}
    \toprule
    Constant                               & Value                  & Role                               & Provenance                                     \\
    \midrule
    \multicolumn{4}{l}{\emph{Spawning} (\mainref{sec:spawn})}                                                                                            \\
    $\theta_0$                             & $2^\circ$              & full-observability angle           & probed ($\times/\div\,2$)                      \\
    $o_0$                                  & $0.1$                  & spawn opacity, unverified          & fixed heuristic                                \\
    $o_1$                                  & $0.3$                  & spawn opacity, fully verified      & probed ($0.2/0.5$)                             \\
    $o_{\mathrm{floor}}$                   & $0.02$                 & opacity without MVS support        & fixed heuristic                                \\
    $\beta_{\mathrm{solid},\mathrm{soft}}$ & $1,5$                  & shape band around init $\beta{=}4$ & structural                                     \\
    $p_{\min}$                             & $1/(0.5 s_{\max})^2$   & coverage floor of sampling         & derived                                        \\
    $\phi$                                 & $2$                    & base sampling scale                & inherited~\cite{meuleman2025onthefly}          \\
    $r_{\mathrm{mvs}}$               & $0.2$                & Agreement                          & fixed heuristic                                \\
    \midrule
    \multicolumn{4}{l}{\emph{Solidity controller} (\mainref{sec:solidity})}                                                                              \\
    $\Delta b$                             & $0.4$                  & per-keyframe shape step            & swept ($0.1$--$0.8$)                           \\
    $o_{\mathrm{solid}}$                   & $0.7$                  & opacity gate for hardening         & probed ($0.5/0.9$)                             \\
    --                                     & $20$\,px, $3$ views    & min.\ owned-pixel support          & fixed heuristic                                \\
    $e_{\mathrm{solid}}$                   & $0.03$                 & mean owned-pixel error band        & fixed heuristic                                \\
    --                                     & $0.5$                  & error consistency for softening    & fixed heuristic                                \\
    --                                     & $10$\,keyframes        & age for ``old''                    & fixed heuristic                                \\
    \midrule
    \multicolumn{4}{l}{\emph{Relocation} (\mainref{sec:error_reloc})}                                                                                    \\
    $o_{\mathrm{dead}}$                    & $0.10$                 & recycling-opacity                  & fixed heuristic                                \\
    --                                     & $10$\,keyframes        & spawn-protection window            & fixed heuristic                                \\
    --                                     & $0.5\times$ mean scale & positional jitter of relocations   & fixed heuristic                                \\
    \midrule
    \multicolumn{4}{l}{\emph{Optimization / rasterization} (\mainref{sec:optim,sec:raster})}                                                     \\
    $K$                                    & $5$                    & spherical-beta colour lobes        & fixed heuristic                                \\
    $\lambda$                              & $0.15$                 & SSIM weight                        & inherited~\cite{meuleman2025onthefly}          \\
    $\lambda_{\mathcal R}$                 & $10^{-2}$              & regularizer weight                 & inherited~\cite{kheradmand2024mcmc,liu2025dbs} \\
    --                                     & $1/255$                & rasterizer contribution cutoff     & convention~\cite{kerbl3dgs}                    \\
    $s_{\max}$                             & $5\%$ of diag.         & max.\ screen-space size            & resolution-relative                            \\
    \midrule
    \multicolumn{4}{l}{\emph{Hierarchy} (\mainref{sec:hierarchy})}                                                                                       \\
    --                                     & $0.5$\,px              & screen-size threshold              & fixed heuristic                                \\
    \bottomrule
  \end{tabular}}
\end{table}

\begin{table}[t]
  \centering\small\setlength{\tabcolsep}{4pt}
  \caption{\textbf{Solidity-controller step sweep on \textsc{Garden}.}
    $\Delta b$ denotes the per-keyframe shape update. ``Solid.'' and ``Soft.'' report the corresponding update counts.}
  \label{tab:suppl-solidity}
  \begin{tabular}{lccrr}
    \toprule
    $\Delta b$ & PSNR$^\uparrow$ & LPIPS$^\downarrow$ & Solid. & Soft. \\
    \midrule
    0.1        & 25.77           & 0.188              & 229k   & 81k   \\
    0.2        & 25.65           & 0.190              & 226k   & 79k   \\
    0.4 (ours) & 25.82           & 0.187              & 224k   & 81k   \\
    0.8        & 25.80           & 0.188              & 220k   & 82k   \\
    \bottomrule
  \end{tabular}
\end{table}

\begin{figure*}[h]
  \centering
  \includegraphics[width=\linewidth]{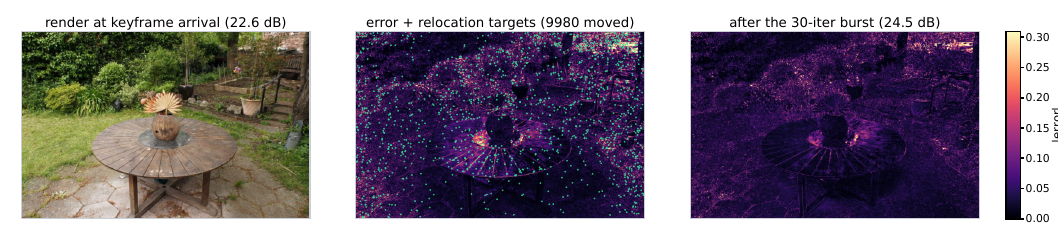}\vspace{-3mm}
  \caption{\textbf{Error-driven relocation} on \textsc{Garden}.
    Left: incoming keyframe and per-pixel error at $22.6$\,dB.
    Middle: dead primitives relocated to high-error contributors, with render-preserving opacity and scale correction.
    Right: the subsequent 30-iteration burst increases quality to $24.5$\,dB.
  }
  \label{fig:suppl-reloc}
\end{figure*}
\begin{table*}[h]
  \centering\small\setlength{\tabcolsep}{4pt}
  \caption{\textbf{One-at-a-time sensitivity study.}
    Each row varies one constant around its default while keeping all other settings fixed.
    Results are averaged over three runs; $\Delta$ reports the PSNR change relative to the default configuration.
  }
  \label{tab:suppl-sensitivity}
  \begin{tabular}{llccc}
    \toprule
    Scene                & Constant                                     & shipped $\to$ probe       & PSNR$^\uparrow$ & $\Delta$                      \\
    \midrule
    \textsc{Garden}      & $o_{\mathrm{solid}}$ (solidify opacity gate) & 0.7 $\to$ 0.5             & 25.78$\pm$0.05  & -0.04\,{\scriptsize$\pm$0.06} \\
    \textsc{Garden}      & $o_{\mathrm{solid}}$ (solidify opacity gate) & 0.7 $\to$ 0.9             & 25.75$\pm$0.08  & -0.07\,{\scriptsize$\pm$0.09} \\
    \textsc{Garden}      & $\theta_0$ (observability angle)             & $2^\circ$ $\to$ $1^\circ$ & 25.78$\pm$0.02  & -0.04\,{\scriptsize$\pm$0.04} \\
    \textsc{Garden}      & $\theta_0$ (observability angle)             & $2^\circ$ $\to$ $4^\circ$ & 25.77$\pm$0.09  & -0.05\,{\scriptsize$\pm$0.10} \\
    \textsc{Garden}      & $o_{\mathrm{1}}$ (verified spawn opacity)    & 0.3 $\to$ 0.2             & 25.83$\pm$0.03  & +0.01\,{\scriptsize$\pm$0.05} \\
    \textsc{Garden}      & $o_{\mathrm{1}}$ (verified spawn opacity)    & 0.3 $\to$ 0.5             & 25.81$\pm$0.08  & -0.01\,{\scriptsize$\pm$0.09} \\
    \midrule
    \textsc{University2} & $\theta_0$ (observability angle)             & $2^\circ$ $\to$ $1^\circ$ & 23.61$\pm$0.04  & -0.06\,{\scriptsize$\pm$0.06} \\
    \textsc{University2} & $\theta_0$ (observability angle)             & $2^\circ$ $\to$ $4^\circ$ & 23.58$\pm$0.06  & -0.09\,{\scriptsize$\pm$0.08} \\
    \bottomrule
  \end{tabular}
\end{table*}

\cref{fig:suppl-reloc} illustrates one relocation step on \textsc{Garden}.
Near-dead primitives are recycled onto persistent high-error regions, keeping the total primitive count unchanged while redistributing capacity.

\cref{tab:suppl-reloc} shows error-based destination
sampling of \mainref{sec:error_reloc} on \textsc{fr2/xyz}.
Both PSNR and LPIPS favor the full method ($28.45$\,dB and $0.133$).
Replacing error weighting with the opacity-based sampling of 3DGS-MCMC~\cite{kheradmand2024mcmc} is worse than disabling relocation ($27.89$\,dB vs.
$28.21$\,dB and $0.148$ vs.
\ $0.136$), indicating that
opacity sampling primarily reinforces already strong regions.

\ifdefined\arxivversion\else\begin{table}[!tb]
  \centering\small\setlength{\tabcolsep}{4pt}
  \caption{\textbf{Unordered input.}
    Identical configuration.
  }
  \label{tab:suppl-shuffle}
  \begin{tabular}{llcccc}
    \toprule
    Scene             & Order    & PSNR$^\uparrow$ & SSIM$^\uparrow$ & LPIPS$^\downarrow$ \\
    \midrule
    \textsc{Garden}   & ordered  & 25.82           & 0.786           & 0.187              \\
                      & shuffled & 25.34           & 0.774           & 0.191              \\
    \midrule
    \textsc{Counter}  & ordered  & 26.98           & 0.860           & 0.222              \\
                      & shuffled & 27.54           & 0.877           & 0.204              \\
    \midrule
    \textsc{Bonsai}   & ordered  & 29.61           & 0.909           & 0.204              \\
                      & shuffled & 27.11           & 0.850           & 0.261              \\
    \midrule
    \textsc{fr1/desk} & ordered  & 23.93           & 0.844           & 0.209              \\
                      & shuffled & 23.09           & 0.870           & 0.179              \\
    \bottomrule
  \end{tabular}
\end{table}
\fi

\section{Constants}
\label{sec:suppl-hparams}

\cref{tab:suppl-hparams} summarizes the constants used by the method and
their provenance.
Several quantities are derived directly from the kernel or image resolution, while the loss weights are inherited from prior work.
The remaining thresholds are chosen from simple heuristics and evaluated with one-at-a-time sensitivity tests in \cref{tab:suppl-sensitivity} and \cref{tab:suppl-solidity}.
Across these probes, quality varies by at most $0.17$\,dB, indicating low sensitivity to the tested operating points.

\section{Full 3DGS Benchmark Table}
\label{sec:suppl-full-benchmark}
See \cref{tab:gs_benchmark_full} for the full benchmark table including GLOMAP with exhaustive mapping, discussed in \mainref{sec:exp-main}.
Sequential mapping fails to register 2 test frames on bicycle and flowers, and 7 on stump.
In \mainref{tab:gs_benchmark}, they are excluded from the average, but they are included in the full table here for completeness as black frames.

\section{Memory Footprint Over Time}
\label{sec:suppl-memory}

\begin{figure}[t]
  \centering
  \IfFileExists{figures/suppl/mem_capacity.pdf}{%
    \includegraphics[width=\linewidth]{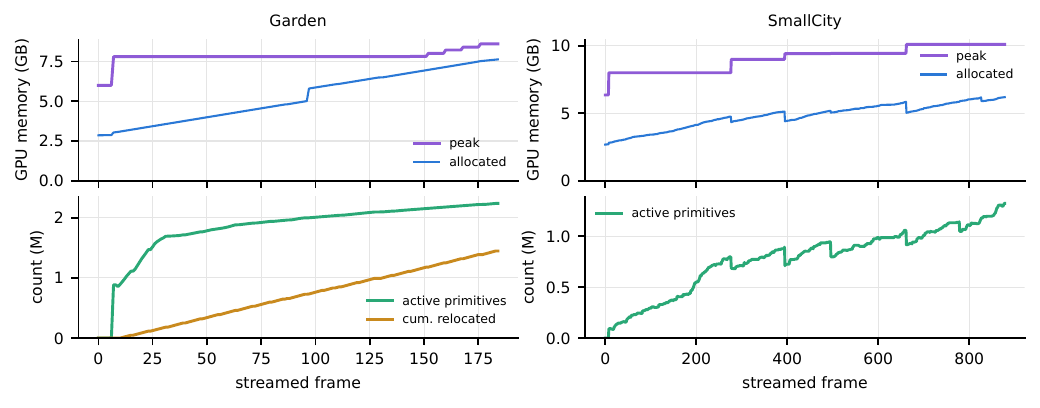}}{%
    \fbox{memory figure pending}}
  \caption{\textbf{Memory and capacity over reconstruction.}
    GPU memory remains bounded while active capacity saturates on \textsc{Garden}.
    Relocation redistributes this capacity over time, whereas hierarchy offloading allows the model to grow on \textsc{SmallCity} without increasing the GPU footprint.
  }
  \label{fig:suppl-memcap}
\end{figure}

The online pass remains well within the 32\,GB GPU budget: peak memory is $8.6$\,GB on \textsc{Garden} (2.2M primitives), $9.3$\,GB on \textsc{fr1/desk}, $6.1$\,GB on \textsc{Truck}, $9.6$\,GB on \textsc{University2} (2.4M), and $10.3$\,GB on \textsc{SmallCity}.
\cref{fig:suppl-memcap} shows that memory saturates while reconstruction
continues.
On \textsc{Garden}, the active count reaches about $2.2$M and changes by only $12\%$ over the second half of the capture, while $1.45$M relocations recycle existing capacity.
On \textsc{SmallCity}, the model grows from $0.85$M to $1.32$M active primitives while peak GPU memory remains near $10$\,GB, with inactive hierarchy levels stored in host memory.

\section{Unordered input}
\label{sec:suppl-unordered}

Our method is designed for unordered image streams, so we evaluate the identical configuration on fixed random permutations of the input images (\cref{tab:suppl-shuffle}).
On \textsc{Garden}, \textsc{Counter}, and \textsc{Bonsai}, shuffling changes PSNR by $-0.48$, $+0.56$, and $-2.50$\,dB, respectively, with comparable registration rates.
It does, however, increase run-to-run variance, e.g.\ $\sigma=0.74$\,dB for shuffled \textsc{Bonsai} versus $0.16$\,dB for the ordered stream.

On \textsc{fr1/desk}, shuffling reduces PSNR by $0.84$\,dB but improves SSIM and LPIPS ($0.844\!
  \to\!0.870$ and $0.209\!\to\!0.179$), with essentially
unchanged registration.
Runtime increases by about $1.5\times$ as temporal adjacency is lost and more frames require shelving and retrieval.
At the same time, random ordering provides a more diverse bootstrap, avoiding the poorly conditioned ordered start observed on this sequence.

Overall, the results confirm that the reconstruction does not require temporal ordering: failed sequential matches fall back to shelf-based bootstrap selection and retrieval over previously registered keyframes.

\section*{LLM Usage Statement}
We used Claude Fable 5.0 and 5.1, Claude Opus 5 and 5.5 (Anthropic), as well as GPT-5.6 Luna (OpenAI) to help draft and edit text and figures, generate and debug code, and develop and check the mathematical formulation in \mainref{sec:hierarchy} (\cref{sec:supp-merge-derivation}).
All outputs were reviewed and verified for accuracy and relevance.

\clearpage  %
\renewcommand{\cmpdir}{figures/figure_comp/figure_comparison_small}

\begin{figure*}[p]
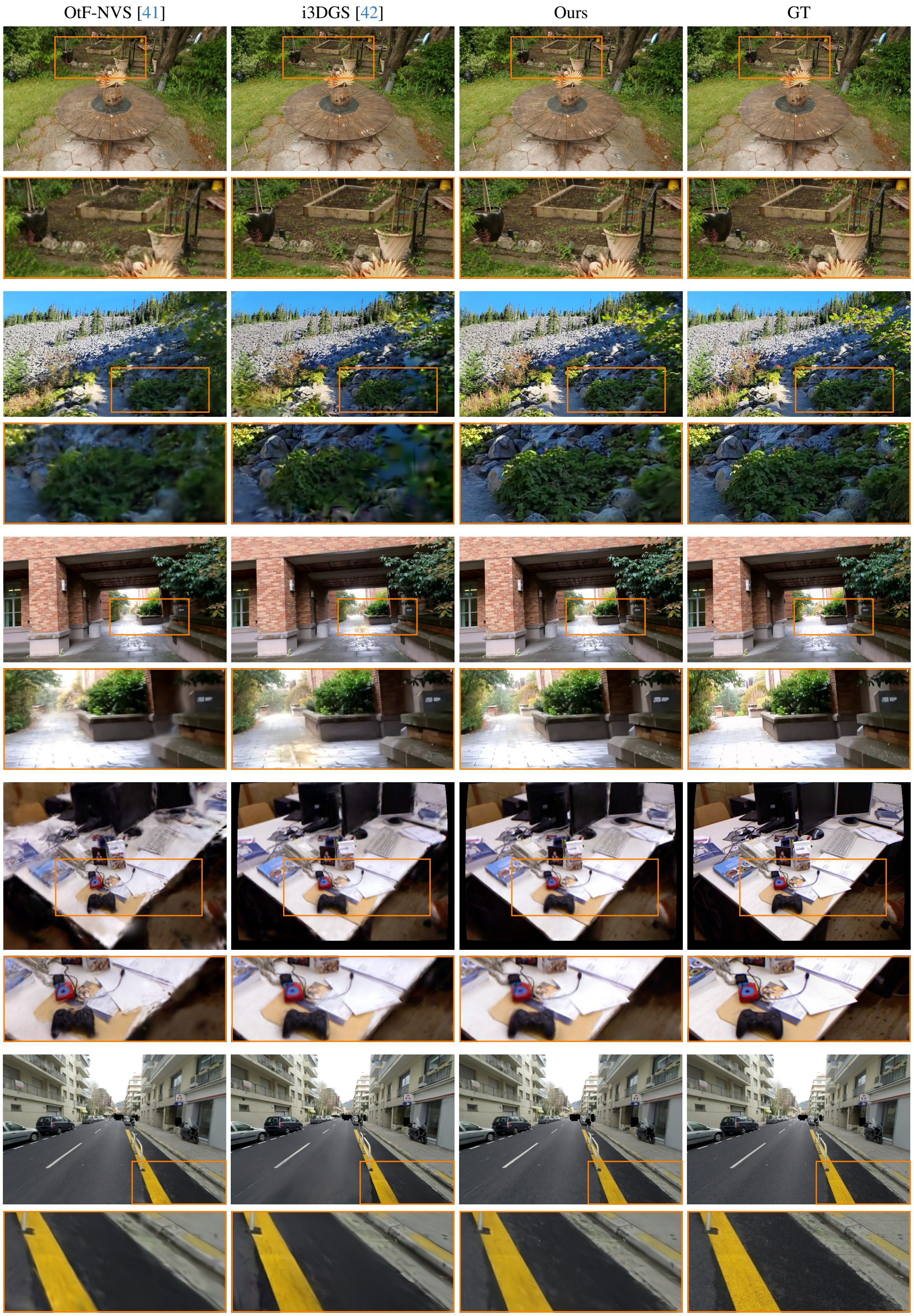

  \centering
  \setlength{\tabcolsep}{1pt}%
  \setlength{\cmpw}{0.205\textwidth}%
  \footnotesize%
  \begin{tabular}{cccc}%
    OtF-NVS~\cite{meuleman2025onthefly} & i3DGS~\cite{meuleman2026immediate} & Ours & GT \\[1pt] \cmpscenefour{full/MipNeRF360_garden} \cmpscenefour{full/StaticHikes_forest1} \cmpscenefour{full/StaticHikes_university2} \cmpscenefour{full/TUM_rgbd_dataset_freiburg1_desk} \cmpscenefour{large_scale/SmallCityFront}\end{tabular} \caption{Qualitative comparison against online reconstruction methods on \textsc{Garden}, \textsc{Forest1}, \textsc{University2}, \textsc{fr1/desk}, and \textsc{SmallCity} (top to bottom).
  }
  \label{fig:comp_online}%
\end{figure*}

\begin{figure*}[p]
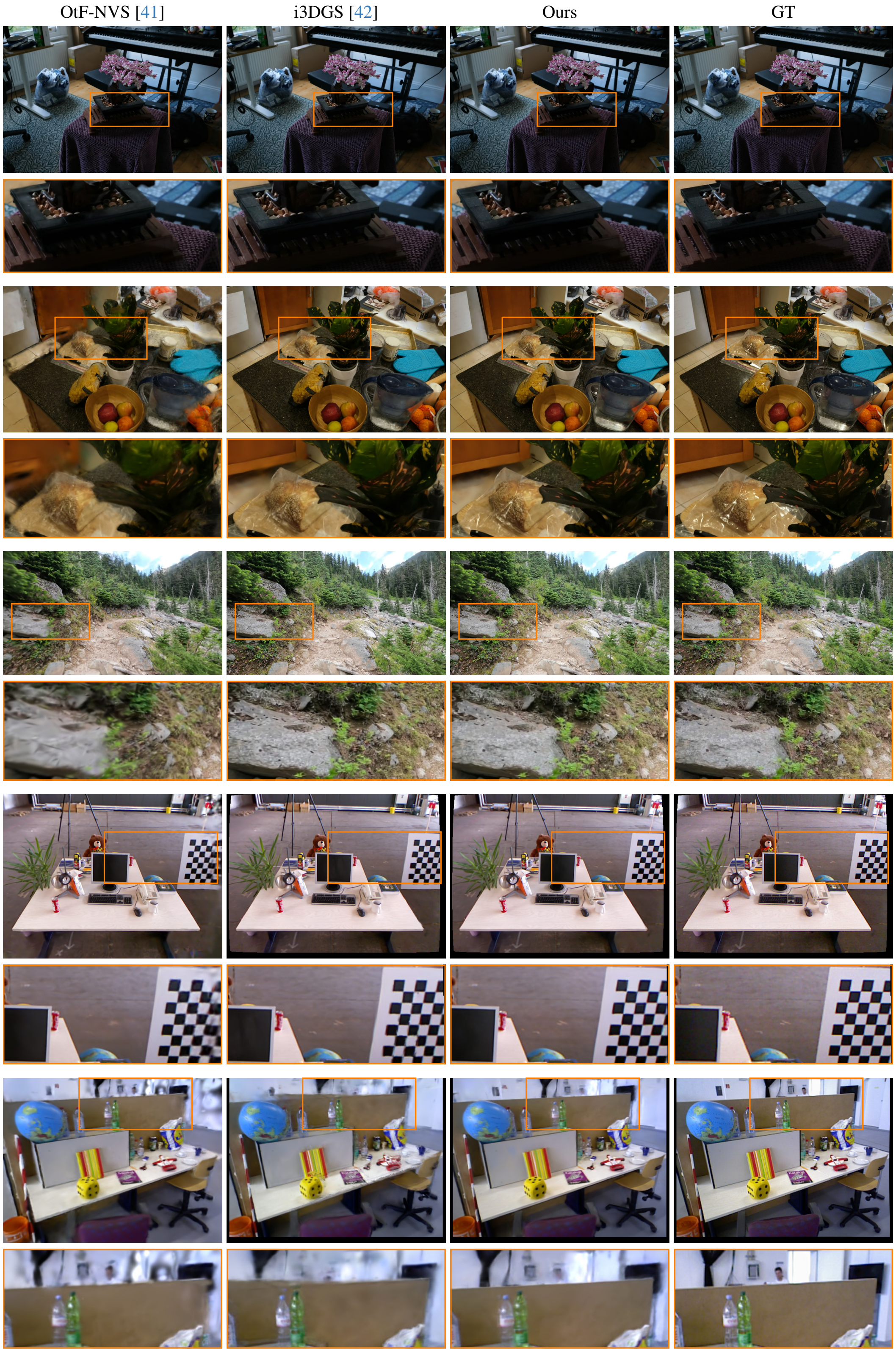

  \centering
  \setlength{\tabcolsep}{1pt}%
  \setlength{\cmpw}{0.196\textwidth}%
  \footnotesize
  \begin{tabular}{cccc}
    OtF-NVS~\cite{meuleman2025onthefly} & i3DGS~\cite{meuleman2026immediate} & Ours & GT \\[1pt] \cmpscenefour{full/MipNeRF360_bonsai} \cmpscenefour{full/MipNeRF360_counter} \cmpscenefour{full/StaticHikes_forest2} \cmpscenefour{full/TUM_rgbd_dataset_freiburg2_xyz} \cmpscenefour{full/TUM_rgbd_dataset_freiburg3_long_office_household}\end{tabular} \caption{Additional qualitative comparisons against online reconstruction methods: \textsc{Bonsai}, \textsc{Counter}, \textsc{Forest2}, \textsc{fr2/xyz}, and \textsc{fr3/office} (top to bottom).
  }
  \label{fig:comp_online_supp}
\end{figure*}

\begin{figure*}[p]
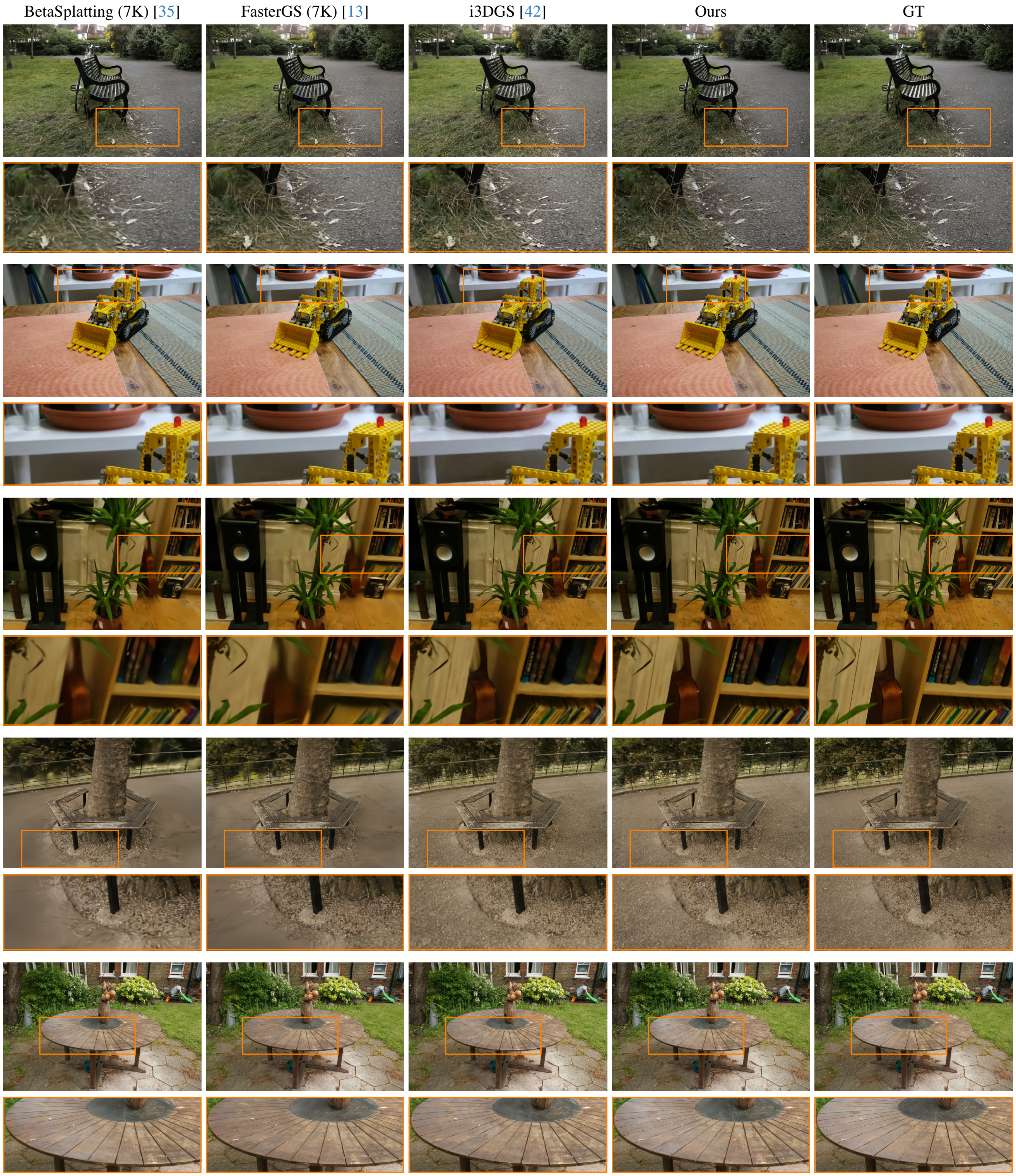

  \centering
  \setlength{\tabcolsep}{1pt}%
  \setlength{\cmpw}{0.194\textwidth}%
  \footnotesize
  \begin{tabular}{ccccc}
    BetaSplatting (7K)~\cite{liu2025dbs} & FasterGS (7K)~\cite{hahlbohm2026fastergs} & i3DGS~\cite{meuleman2026immediate} & Ours & GT \\[1pt] \cmpscenefive{mip_tandt_db/MipNeRF360U_bicycle} \cmpscenefive{mip_tandt_db/MipNeRF360U_kitchen} \cmpscenefive{mip_tandt_db/MipNeRF360U_room} \cmpscenefive{mip_tandt_db/MipNeRF360U_bonus_treehill} \cmpscenefive{mip_tandt_db/MipNeRF360_garden}\end{tabular} \caption{Additional qualitative comparisons against offline baselines: \textsc{Bicycle}, \textsc{Kitchen}, \textsc{Room}, \textsc{Treehill}, and \textsc{Garden} (top to bottom).
  }
  \label{fig:comp_offline_supp_a}
\end{figure*}

\begin{figure*}[p]
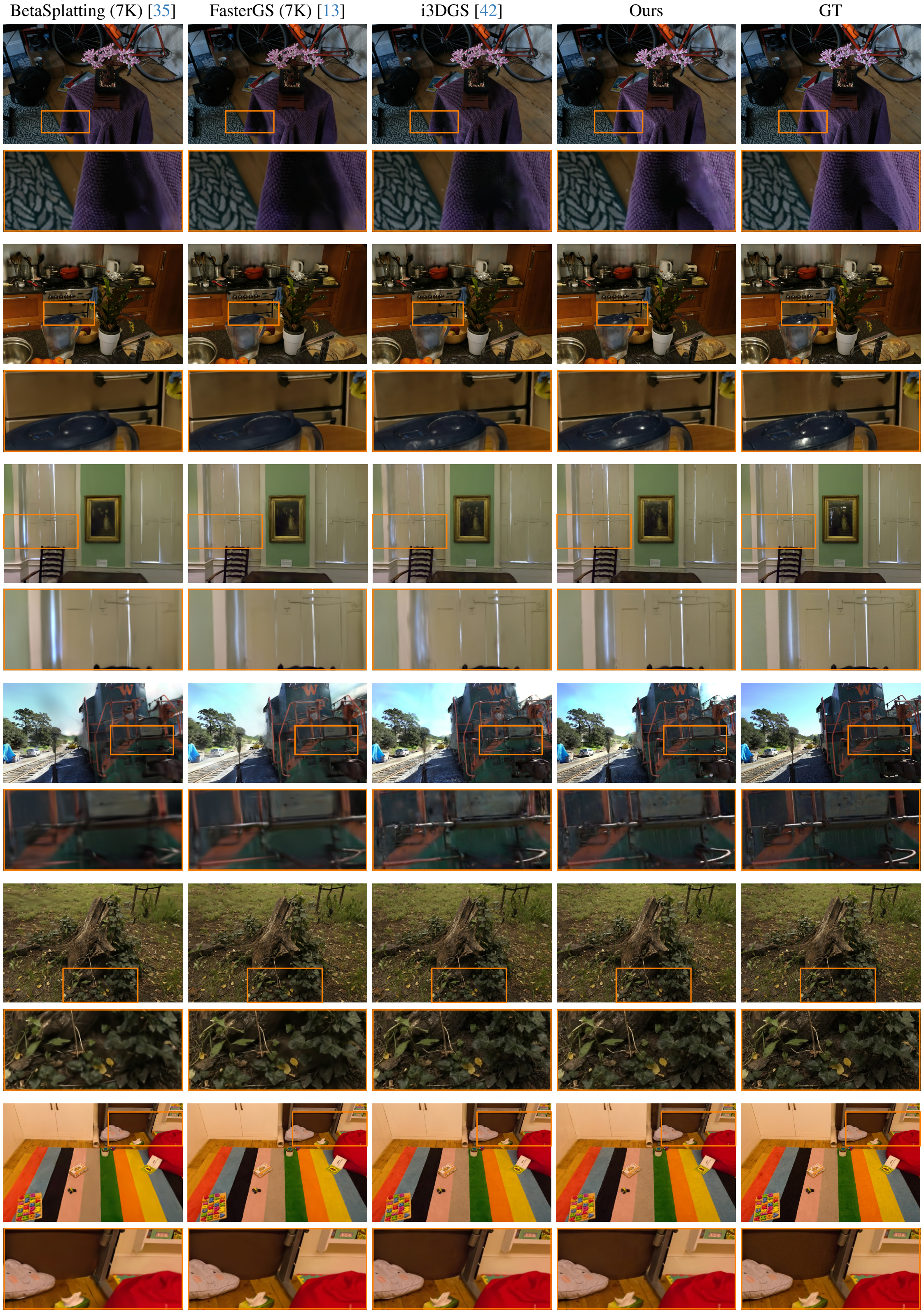

  \centering
  \setlength{\tabcolsep}{1pt}%
  \setlength{\cmpw}{0.165\textwidth}%
  \footnotesize
  \begin{tabular}{ccccc}
    BetaSplatting (7K)~\cite{liu2025dbs} & FasterGS (7K)~\cite{hahlbohm2026fastergs} & i3DGS~\cite{meuleman2026immediate} & Ours & GT \\[1pt] \cmpscenefive{mip_tandt_db/MipNeRF360_bonsai} \cmpscenefive{mip_tandt_db/MipNeRF360_counter} \cmpscenefive{mip_tandt_db/db_drjohnson}\cmpscenefive{mip_tandt_db/tandt_train} \cmpscenefive{mip_tandt_db/MipNeRF360U_stump} \cmpscenefive{mip_tandt_db/db_playroom}\end{tabular} \caption{Additional qualitative comparisons against offline baselines: \textsc{Bonsai}, \textsc{Counter}, \textsc{Dr.~Johnson}, \textsc{Train}, \textsc{Stump}, and
    \textsc{Playroom} (top to bottom).}
  \label{fig:comp_offline_supp_b}
\end{figure*}

\end{document}